\documentclass[11pt]{article}
\usepackage{acl}
\usepackage{amsmath}
\usepackage{times}
\usepackage{latexsym}
\usepackage{graphicx}
\usepackage{tcolorbox}
\usepackage{lipsum} 
\usepackage{hyperref}

\usepackage[table,xcdraw]{xcolor}
\usepackage[T1]{fontenc}
\usepackage{algorithm}
\usepackage[utf8]{inputenc}
\usepackage{algpseudocode}
\usepackage{microtype}

\usepackage{inconsolata}
\usepackage{enumitem}

\usepackage{graphicx}

\usepackage[utf8]{inputenc} 
\usepackage[T1]{fontenc}    
\usepackage{url}            
\usepackage{booktabs}       
\usepackage{amsfonts}       
\usepackage{nicefrac}       
\usepackage{microtype}      
\usepackage{multirow}
\usepackage{graphicx}
\usepackage{pifont}
\usepackage{makecell}
\usepackage{enumitem}
\usepackage{array}
\usepackage{colortbl}
\usepackage{amssymb}  
\usepackage{bbding}    
\usepackage{wasysym}   
\definecolor{darkgreen}{RGB}{0,100,0}
\definecolor{mia-green}{RGB}{78, 205, 196}
\definecolor{qwen-red}{RGB}{255, 107, 107}
\definecolor{mia-rag-green}{RGB}{39, 174, 96}
\definecolor{baseline-gray}{RGB}{149, 165, 166}
\definecolor{darkblue}{rgb}{0, 0, 0.5}
\hypersetup{colorlinks=true, citecolor=darkblue, linkcolor=darkblue, urlcolor=darkblue}

\newcounter{rqsection}
\renewcommand{\therqsection}{(\emph{H\arabic{rqsection}})}

\newcommand{\rqsection}[1]{
  \refstepcounter{rqsection} 
  \medskip
  \noindent{\therqsection\, \emph{#1}} 
}

\newcounter{rqsous}[rqsection]
\renewcommand{\therqsous}{(\emph{H\arabic{rqsection}.\arabic{rqsous}})}

\newcommand{\rqsous}[1]{%
  \refstepcounter{rqsous}
  \medskip
  \noindent{\therqsous\, \emph{#1}}%
}

\title{Mindscape-Aware Retrieval Augmented Generation\\ for Improved Long Context Understanding}
\author{
Yuqing Li$^{1,2}$\thanks{\,\,Equal contribution \quad $^{\dagger}$ Corresponding authors} { }
Jiangnan Li$^{3*}${ }
Zheng Lin$^{1,2\dagger}${ } 
Ziyan Zhou$^{1,2}$  \\
\textbf{Junjie Wu$^{4}${ } 
Jie Zhou$^{3}$ { }Mo Yu$^{3\dagger}$} \\
$^{1}$Institute of Information Engineering, Chinese Academy of Sciences \\
$^{2}$School of Cyber Security, University of Chinese Academy of Sciences \\
$^{3}$WeChat AI, Tencent \quad
$^{4}$Hong Kong University of Science and Technology \\
\texttt{\{liyuqing, linzheng\}@iie.ac.cn}\quad
\texttt{\{jiangnanli, moyumyu\}@tencent.com}
}

\begin{document}
\maketitle

\begin{abstract}

Humans understand long and complex texts by relying on a holistic semantic representation of the content. This global view helps organize prior knowledge, interpret new information, and integrate evidence dispersed across a document, as revealed by the \textit{Mindscape-Aware Capability} of humans in psychology.
However, current Retrieval-Augmented Generation (RAG) systems often lack explicit global semantic guidance, making it difficult to retrieve and integrate dispersed evidence in long-context tasks.
In this paper, we propose Mindscape-Aware RAG (MiA-RAG), the first framework to formulate mindscape-aware retrieval and generation as a unified conditioning paradigm for LLM-based RAG.
MiA-RAG builds a mindscape through hierarchical summarization and conditions both retrieval and generation on this global semantic representation. This enables the retriever to form enriched query embeddings and the generator to reason over retrieved evidence within a coherent global context.
We evaluate MiA-RAG across diverse long-context and bilingual benchmarks for evidence-based understanding and global sense-making. It consistently surpasses baselines, and further analysis shows that it aligns local details with a coherent global representation, enabling more human-like long-context retrieval and reasoning. The accompanying materials are available at \url{https://huggingface.co/MindscapeRAG/MiA-Emb-8B}.
\end{abstract}

\begin{figure}
    \centering
    \includegraphics[width=0.94\linewidth]{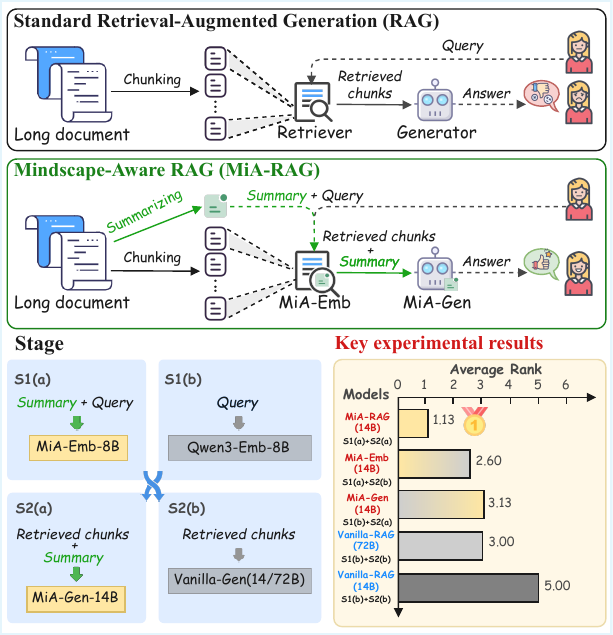}
    \caption{Average model ranks across five long-context benchmarks under 3/5/10-chunk settings.}
\label{fig:teaser}
\end{figure}
\section{Introduction}

Human thinking is inherently context-dependent.
For any learned topic, familiar situation, or ongoing project of engagement, humans maintain a global semantic representation in memory.
When the same topic reappears, this global memory is reactivated, endowing humans with the \textbf{Mindscape-Aware capability} to become aware of the approximate scope of their knowledge and to rely on this memory to \emph{interpret new inputs within context, selectively channel retrieval toward context-relevant knowledge, and guide subsequent reasoning accordingly}.
This phenomenon is grounded in theories from psychology~\cite{bartlett1932remembering,tulving1973encoding,reyna1995fuzzy} and neuroscience~\cite{ralph2017neural}, which posit that when a topic is reactivated, conscious cognition is constrained and guided by globally integrated knowledge, with converging support from neuroimaging observations. We review both the theoretical and empirical supports in Appx.~\ref{sec:supports}.

Retrieval-Augmented Generation (RAG)~\cite{zhu2025,gao2023retrieval,zhang2025survey} has emerged as a critical strategy for long-context understanding by retrieving useful context fragments from very long inputs, thereby overcoming LLMs’ limited context lengths~\cite{lewis2020retrieval}.
However, current RAG systems primarily retrieve and generate based on local, evidence-level signals, lacking the \emph{mindscape-aware} capability to activate a global semantic frame as humans do.
Endowing RAG with this capability is therefore especially promising for personalized knowledge collections, such as long-context question answering~\cite{bai2023longbench}, code generation~\cite{wang2024coderag}, and AI assistants over personal projects~\cite{notebookLM}.
Specifically, the cognitive advantages of the mindscape translate naturally into the following benefits:

\begin{itemize}[noitemsep,nolistsep,leftmargin=*]
    \item \emph{\textbf{Enriched Understanding}}: supported by awareness of global semantics that fills missing information and resolves underspecified meanings.
    \item \emph{\textbf{Selective Retrieval}}: biases query embeddings toward the active topic's conceptual frame, filtering out ambiguities arising from other topics.
    \item \emph{\textbf{Integrative Reasoning}}: interprets retrieved results within the global context to ensure coherent synthesis and understanding.
\end{itemize}

Motivated by these insights, we propose the first approach to equip LLM-based RAG systems with mindscape-aware capabilities.
Specifically, we tackle the long-context understanding problem by approximating the global impression of a long document with a summary generated in a hierarchical manner, which serves as an external representation of global memory.
Taking this form of global memory as an additional input, we train models to fit two core functions of our new \emph{\textbf{Mindscape-Aware RAG}} (\textbf{MiA-RAG}) framework:
\begin{itemize}[noitemsep,nolistsep,leftmargin=*]
\item \emph{\textbf{Mindscape-Aware Retrieval }}
The mindscape-aware capability enables queries to be understood within their global semantic context, producing query representations that are not only anchored to the topical scope (\emph{Selective Retrieval}), but also integrated with global contextual information (\emph{Enriched Understanding}).
We instantiate these capabilities through a specially trained embedding model, which takes both the global memory and the query as input, and learns to integrate global information into query embeddings to enhance retrieval performance.

\item \emph{\textbf{Mindscape-Aware Generation }}
Solely enhancing the RAG pipeline with contextually informed query embeddings introduces a new challenge: the generator becomes weaker than the retriever, as it lacks access to the global context.
Consequently, the generator may misinterpret the relevance of the retrieved information or fail to effectively utilize it, even when the retrieved content contains the correct evidence.
To mitigate this asymmetry, we condition generation on the same global memory. By incorporating the global memory into the generator's inputs, the model learns to interpret the retrieved chunks and their relationship to the query within the broader global context (\emph{Integrative Reasoning}), leading to more faithful reasoning and answers.
\end{itemize}

We evaluate MiA-RAG across a range of long-context understanding tasks spanning diverse domains and genres (\emph{e.g.}, government reports, narratives), in both English and Chinese.
The evaluation also covers various task formats, including free-form QA, multiple-choice QA, and claim verification, as well as different RAG configurations such as vanilla RAG and GraphRAG~\cite{edge2024local}.
As summarized in Figure~\ref{fig:teaser}, the MiA family is more effective than baselines; in particular, MiA-RAG-14B achieves the best average rank, surpassing the vanilla 72B system and highlighting the benefit of mindscape-aware retrieval and generation.

Beyond performance gains, we analyze MiA-RAG’s internal mechanisms via embedding-space geometry and a new mindscape-coherent metric, showing that it internalizes global semantics: the mindscape reshapes query representations toward the global semantic space and acts as a scaffold that guides attention for Integrative Reasoning.

Our contributions are summarized as follows:

(1) 
We formulate the psycho- and neuro-inspired problem of mindscape-aware thinking, and present the first computational solution that equips LLMs with this capability.

(2)
We conduct extensive experiments under diverse settings, demonstrating the necessity and effectiveness of integrating mindscape-aware capability into LLM-based systems.

(3)
Our in-depth analysis reveals that the mindscape aligns the geometry of query representations with the global semantic space and serves as a semantic scaffold to guide attention, confirming the active internalization of global context rather than surface-level pattern matching.

\begin{figure*}
    \centering
    \includegraphics[width=0.93\linewidth]{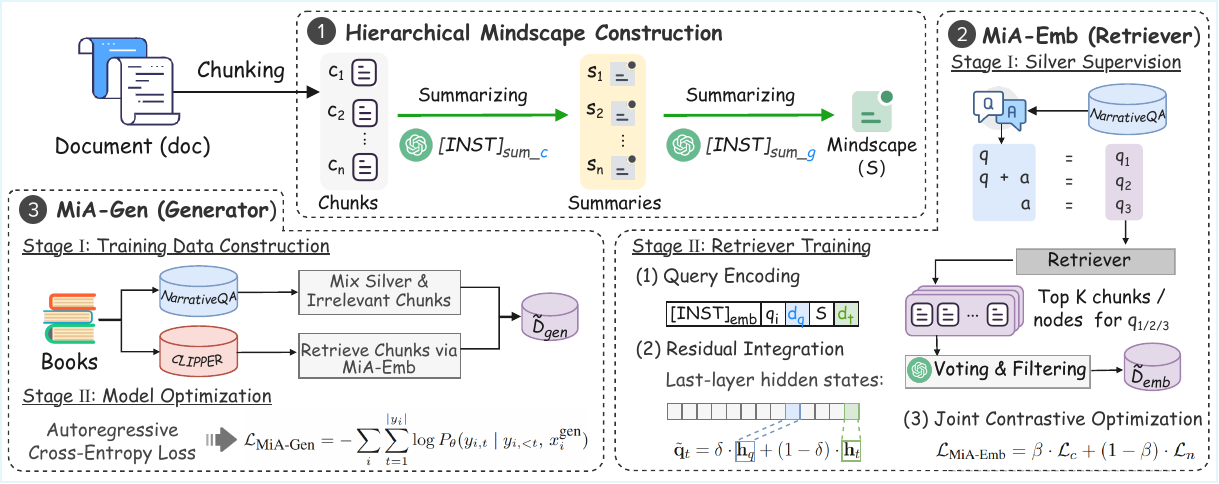}
    \caption{Overview of the framework of MiA-RAG.}
    \label{fig:framework}
\end{figure*}
\section{Related Work}
\paragraph{Context-Aware Embeddings}
Our MiA-Emb is related to the research topic of context-aware retrieval~\cite{anthropic}.
This line of work mainly focuses on producing embeddings enriched with contextual information.
A straightforward approach is to encode each chunk within a long-context window using LLMs that support extended inputs~\cite{bge-m3,sturua2024jinaembeddingsv3multilingualembeddingstask,nussbaum2024nomic,wang2024improving,li2023towards,voyage2025context3}.
To incorporate information beyond the local window, \citet{xu2024fine} construct graphs of discourse relations and then utilize graph neighbors for augmentation.
While these methods provide additional context, they do not directly teach the model how to fuse such information. 
Recently, \citet{wu2025sitemb} introduce training techniques that enable the embedding model to more effectively situate a chunk within its local neighborhood, achieving state-of-the-art retrieval performance with only modest training data requirements. 
However, it primarily enhances retrieval by enriching the chunk representation with local context. 
In contrast, our method injects global semantics into the query representation, guiding queries toward the correct semantic region of the index and improving selective retrieval without modifying the original chunk embeddings.

\paragraph{Long Context Compression}
To capture the overarching semantics of long documents, recent work moves beyond token retention or KV-cache compression~\cite{yang2024pyramidinfer,li2024snapkv,xiao2023efficient} toward abstractive compression and global memory~\cite{qian2025memorag,behrouz2024titans}.
\text{MemoRAG}~\cite{qian2025memorag} uses compressed memory for explicit planning, while \text{LeanRAG}~\cite{zhang2026leanrag} organizes knowledge into hierarchical semantic graphs for structure-guided retrieval.
In contrast,  MiA-RAG learns to internalize the mindscape as a shared conditioning signal for both retrieval and generation, avoiding intermediate search steps or external graph construction.
\citet{behrouz2024titans} learn test-time long-term memory, whereas our mindscape is a persistent global semantic anchor constructed from the full text.

\section{Method}

In this section, we introduce the Mindscape-Aware RAG (MiA-RAG) framework (Figure~\ref{fig:framework}), which emulates how humans leverage global context to support understanding, retrieval, and reasoning.

\subsection{Preliminaries}
Let $doc$ be a long document that has been partitioned into chunks ${c_i \in C}$.
In a vanilla RAG pipeline, a retriever selects a set of chunks $C_{\text{ret}} \subseteq C$ for a query $q$, and the generator conditions on these chunks to produce an answer $a$:
\begin{equation}
    q \xrightarrow{C_{\text{ret}}} a.
\end{equation}

However, this setup does not provide a global view of the document for either retrieval or generation.
To bridge this gap, we propose MiA-RAG, which incorporates an explicit global semantic scaffold termed the Mindscape $S$. By conditioning both retrieval and generation on $S$, MiA-RAG situates local evidence within a global context, improving retrieval accuracy and reasoning consistency:
\begin{equation}
  {q,S} \xrightarrow{{C_{ret},S}} a.
\end{equation}

\subsection{Hierarchical Mindscape Construction}
\label{sec:summary_gen}

We construct the Mindscape $S$ via bottom-up hierarchical summarization.

We first prompt  summarizer $\mathcal{M}_s$ (GPT-4o) with $\textcolor{gray}{\texttt{[INST]}_{\texttt{sum\_c}}}$  (Appx.~\ref{appendix:prompt}) to summarize each chunk:
\begin{equation}
    s_i = \mathcal{M}_s(\textcolor{gray}{\texttt{[INST]}_{\texttt{sum\_c}}}, c_i).
\end{equation}

After obtaining the chunk-level summaries, the sequence of $\{s_i\}$ is concatenated in order and further summarized using $\textcolor{gray}{\texttt{[INST]}_{\texttt{sum\_g}}}$ (Appx.~\ref{appendix:prompt}) to produce a single global representation:
\begin{equation}
    S = \mathcal{M}_s(\textcolor{gray}{\texttt{[INST]}_{\texttt{sum\_g}}},\, [s_1, s_2, \ldots, s_n]).
\end{equation}

The resulting $S$ serves as a document-level summary and can be updated incrementally by regenerating only the affected paths from modified chunks to the root~\cite{chang2023booookscore}.

\subsection{MiA-Emb: Mindscape-Aware Retriever}
To obtain the  Mindscape-Aware Retriever (MiA-Emb), we construct a supervision dataset and optimize the model with a multi-task objective.

\label{sec:mia_retriever}
\subsubsection{Supervision Construction}
\label{sec:silver_chunk}

Existing long-narrative understanding datasets such as NarrativeQA~\cite{kovcisky2018narrativeqa} typically provide only QA pairs and do not link questions to fine-grained supporting evidence. Such supervision is essential for training long-context retrievers, whether evidence is represented as text chunks in standard RAG or as graph nodes in GraphRAG.

Given the cost of manual annotation, we automatically extend NarrativeQA to curate $\tilde{D}_\text{emb}$, a dataset offering silver-standard alignments at both chunk and node levels.  For chunk evidence, we perform \textit{query augmentation}, \textit{majority-vote ensemble} retrieval, and \textit{LLM-based filtering} to identify silver chunks (Algorithm~\ref{alg:silver_evidence} in Appx.~\ref{appendix:silver_annotate}). As shown in Table~\ref{tab:silver_chunk_annotate}, an oracle experiment validates these annotations: using only silver chunks ($\sim$1/30 tokens) already exceeds full-context performance. Node evidence is constructed in an analogous way.

In total, $\tilde{D}_\text{emb}$ comprises 27{,}117 questions, averaging 2.3 silver chunks and 2.9 nodes per question. Details of the construction are provided in Appx.~\ref{appendix:silver_annotate}.

\begin{table}[ht!]
\scriptsize
\centering
\begin{tabular}{lll|cc}
\toprule
\textbf{Input}&\textbf{\# Books}& \textbf{Avg. Chunks} & \textbf{EM} & \textbf{F1} \\ 
\midrule 
Silver Chunks &20 &2.3  &\textbf{31.29}  &\textbf{52.91} \\ 
Oracle (Full Book) &20 &69.2  &30.04  &50.79 \\ 
\bottomrule
\end{tabular}
\caption{Validation of silver chunk annotations.}
\label{tab:silver_chunk_annotate}
\end{table}
\subsubsection{Model Optimization}
\label{sec:mia_emb}
We develop MiA-Emb to explicitly inject global context into query representations.
Specifically, we encode each query $q_i$ by explicitly incorporating the summary $S$ together with task-specific control tokens $d$. The input sequence is defined as
\begin{equation}
Q = [\textcolor{gray}{\texttt{[INST]}_{\texttt{emb}}};\ q_i;\ \textcolor{gray}{d_q};\ S;\ \textcolor{gray}{d_t}],
\end{equation}
where $\textcolor{gray}{\texttt{[INST]}_{\texttt{emb}}}$ (Appx.~\ref{appendix:prompt}) is the  instruction, $\textcolor{gray}{d_q}$ marks the end of query, and $\textcolor{gray}{d_t}=[\textcolor{gray}{d_n}, \textcolor{gray}{d_c}]$ activates node- and chunk-retrieval modes, respectively.

To balance the query intent with global guidance, we use a residual integration and train the model using an InfoNCE objective~\cite{Oord2018RepresentationLW}. Full details are provided in Appx.~\ref{appendix:mia_emb_opt}.

\subsection{MiA-Gen: Mindscape-Aware Generator}
\subsubsection{Training Data Construction}
We develop the Mindscape-Aware Generator (\textbf{MiA-Gen}) by fine-tuning on a composite corpus derived from the training splits of two datasets: \textbf{NarrativeQA} for long context QA and \textbf{CLIPPER}~\cite{pham2025clipper}, a synthetic dataset featuring narrative claims for verification.
Each training instance is formatted as follows:
\begin{equation}
\underbrace{
\langle \texttt{[INST]}_{\texttt{gen}}\;;\; S;\; \hat{C}_{\text{ret}};\; q_i \rangle
}_{x_i^{\text{gen}}}
\;\rightarrow\; y_i ,
\end{equation}
where  $\hat{C}_{\text{ret}}$ is the retrieved chunks, and $y_i$ is the target. Instruction $\texttt{[INST]}_{\texttt{gen}}$ is detailed in Appx.~\ref{appendix:prompt}.

To ensure robustness against retrieval noise, we construct $\hat{C}_{\text{ret}}$ to simulate realistic inference conditions. For NarrativeQA, we mix silver evidence chunks with irrelevant text; for CLIPPER, we utilize the actual top-$k$ retrieval results from MiA-Emb. Furthermore, we augment the data by varying the number of chunks in $\hat{C}_{\text{ret}}$ ($k \in \{3, 5, 10\}$) for each query, enabling the model to handle diverse context lengths.
The aggregation of these instances constitutes the final dataset for the generator $\tilde{\mathcal{D}}_{\text{gen}}$.




\subsubsection{Model Optimization}
MiA-Gen is optimized over  $\tilde{\mathcal{D}}_{\text{gen}}$ using the autoregressive cross-entropy loss:
\begin{equation}
\mathcal{L}_{\text{MiA-Gen}} = - \sum_i \sum_{t=1}^{|y_i|} \log P_{\theta}(y_{i,t} \mid y_{i,<t},\, x_i^{\text{gen}}). 
\end{equation}

\begin{table*}[t]
\centering
\small
\renewcommand{\arraystretch}{1.0}
\resizebox{\textwidth}{!}{
\begin{tabular}{m{2.5cm}|
                m{2.6cm}m{0.3cm}|
                m{1.9cm}m{0.3cm}|
                c|c|c|c|c|c}
\toprule
\multirow{2}{*}{\textbf{Model}} &
\multicolumn{2}{c|}{\textbf{Retriever}} &
\multicolumn{2}{c|}{\textbf{Generator}} &
\textbf{NarrativeQA} &
\textbf{$\infty$ Bench} &
\textbf{Det.QA-Zh} &
\textbf{Det.QA-En} &
\textbf{NoCha} &
\multirow{2}{*}{\textbf{Avg.}} \\
\cmidrule(lr){2-3} \cmidrule(lr){4-5}
& \textbf{Emb. Model} & \textbf{+S} &
  \textbf{Gen. Model} & \textbf{+S} &
  F1 & Acc & Acc & Acc & Pair Acc & \\
\midrule

Summary-Only & \textbf{-} &-& Qwen2.5-72B &\ding{51}
& 39.24 & 72.05 & 73.67 & 61.33 & 31.75 & 55.61 \\

Vanilla-RAG & {Qwen3-Emb-8B} &\ding{55}& Qwen2.5-72B &\ding{55}
& 41.13/45.51/49.06
& 75.55/80.79/86.90
& 63.67/70.83/78.00
& 55.50/61.33/71.17
& 33.33/38.10/41.27
& 59.48 \\

HippoRAG-v2 & Qwen3-Emb-8B & \ding{55} & Qwen2.5-72B & \ding{55} &40.06/45.20/47.16  & 79.48/84.28/85.15 &70.33/73.83/77.17 &64.83/68.00/68.50 &33.33/36.51/47.62 &61.43
\\

MemoRAG &BGE-M3 &\ding{55}&Qwen2.5-72B &\ding{55}&40.72/43.63/44.86 & 75.11/76.42/78.17 &67.50/68.83/71.17 &  61.17/63.50/64.33&\textbf{46.03}/\textbf{49.21}/49.21 &59.99\\

Raptor &Qwen3-Emb-8B  &\ding{55}& Qwen2.5-72B &\ding{55} & 39.48/42.76/46.94 & 75.98/79.91/83.84  & 62.00/68.67/74.67 & 54.50/62.00/69.67 & 34.92/34.92/46.03 & 58.42\\

\rowcolor[gray]{0.9}MiA (Gen-Only) & {Qwen3-Emb-8B}&\ding{55} & Qwen2.5-72B&\ding{51}
& 47.67/48.46/51.81
& 82.10/83.84/86.46
& 76.00/78.50/81.33
& 68.17/69.67/73.33
& 36.51/42.86/44.44
& 64.74 \\

\rowcolor[gray]{0.9}MiA (Emb-Only) & {MiA-Emb-8B}&\ding{51} & Qwen2.5-72B&\ding{55}
& 46.38/48.06/49.88
& \textbf{84.72/87.77/90.39}
& 76.17/81.17/82.67
& 67.17/71.83/75.33
& \text{42.86}/42.86/49.21
& 66.43 \\

\rowcolor[gray]{0.75}MiA & {MiA-Emb-8B} &\ding{51}& Qwen2.5-72B&\ding{51}
& \textbf{50.05/51.04/53.15}
& {84.71/86.46/88.21}
& \textbf{81.67/83.17/84.17}
& \textbf{70.33/72.33/75.50}
& {41.27}/44.44\textbf{/52.38}
& \textbf{67.93} \\

\midrule
\midrule

Summary-Only & \textbf{-} &- & Qwen2.5-14B&\ding{51}
& 38.03 & 61.57 & 70.50 & 58.00 & 17.46 & 49.11 \\

Vanilla-RAG & {Qwen3-Emb-8B} &\ding{55}& Qwen2.5-14B&\ding{55}
& 39.32/40.99/44.38
& 72.49/73.80/77.29
& 60.33/68.50/75.67
& 55.17/59.83/66.67
& 17.46/11.11/15.87
& 51.93 \\

HippoRAG-v2 & Qwen3-Emb-8B & \ding{55} & Qwen2.5-14B & \ding{55} & 38.34/41.28/43.60 &76.86/79.04/80.79 & 65.67/71.00/75.00 &58.00/61.33/65.83 &20.63/15.87/17.46 &54.05\\

MemoRAG &BGE-M3 &\ding{55}& Qwen2.5-14B &\ding{55} &36.58/42.87/44.42 &71.18/72.05/73.80  &60.70/66.17/68.83 &54.33/55.67/58.50 &28.57/31.75/34.92 &53.35\\
Raptor &Qwen3-Emb-8B  &\ding{55}& Qwen2.5-14B &\ding{55} & 36.81/39.92/44.00 &75.55/82.10/82.53 &60.50/63.83/71.33 & 53.50/60.00/65.50 &  22.22/14.29/15.87 & 52.53\\

\rowcolor[gray]{0.9}{MiA (Gen-Only)} & {Qwen3-Emb-8B} &\ding{55}& Qwen2.5-14B&\ding{51}
& 43.04/43.32/45.83
& 75.98/80.79/79.48
& 75.33/78.83/80.17
& 65.33/66.83/70.00
& 19.05/15.87/26.98
& 57.79 \\

\rowcolor[gray]{0.9}{MiA (Gen-Only)} & {Qwen3-Emb-8B}&\ding{55} & MiA-Gen-14B&\ding{51}
& 50.55/50.08/51.99
& 75.98/82.10/80.79
& 76.17/78.50/79.50
& 67.67/\textbf{71.50}/71.67
& 49.21/47.62/50.79
& 65.61 \\

\rowcolor[gray]{0.9}MiA (Emb-Only) & {MiA-Emb-8B}&\ding{51} & Qwen2.5-14B&\ding{55}
& 45.89/46.77/47.13
& 79.48/82.97/84.28
& 73.00/77.17/80.83
& 62.33/65.83/70.33
& 22.22/26.98/26.98
& 59.48 \\
\rowcolor[gray]{0.75}MiA & {MiA-Emb-8B}&\ding{51} & {Qwen2.5-14B}&\ding{51}
& 44.38/46.66/47.87
& 79.91/\textbf{83.41}/84.28
& 78.33/79.00/\textbf{82.00}
& 66.50/69.17/71.82
& 30.16/28.57/38.10
& 62.01 \\

\rowcolor[gray]{0.75}MiA-RAG & {MiA-Emb-8B}&\ding{51} & {MiA-Gen-14B}&\ding{51}
& \textbf{52.48/53.52/53.56}
&\textbf{80.79}/81.22/\textbf{85.15}
& \textbf{79.00/79.67}/81.83
& \textbf{69.00}/71.17/\textbf{75.50}
& \textbf{55.56/49.21/53.97}
& \textbf{68.11} \\

\bottomrule
\end{tabular}
}
\caption{Overall results on Long-story QA and Reasoning benchmarks under top-3/5/10 retrieval. “+S” indicates whether the summary is incorporated at that stage.
Darker gray rows represent deeper mindscape involvement; bold marks the best per scale. The final column reports the average over all metrics per row. Notably, MemoRAG~\cite{qian2025memorag} uses an additional 7B memory model to enrich retrieval, as detailed in Appx.~\ref{appendix:baselines}.
}
\label{tab:qa_overall_avg}
\end{table*}

\section{Experimental Setting}

\paragraph{Evaluated Models}
MiA-RAG is implemented by fine-tuning Qwen-series~\cite{zhang2025qwen3,qwen2.5} models. We develop two components:
\begin{itemize}[noitemsep, topsep=2pt, leftmargin=*]
    \item \textbf{Mindscape-Aware Retriever}: we fine-tune \textit{Qwen3-Embedding-8B} to obtain MiA-Emb.
    \item \textbf{Mindscape-Aware Generator}: we fully fine-tune \textit{Qwen2.5-14B-Instruct} to obtain MiA-Gen.
\end{itemize}
We compare against Qwen baselines and state-of-the-art RAG methods, including HippoRAG v2~\cite{guti'errez2025}, MemoRAG~\cite{qian2025memorag} and RAPTOR~\cite{raptor}.

\paragraph{Public Long Narrative Understanding Tasks}
We evaluate on long-narrative understanding benchmarks, many of which exceed common LLM input limits (e.g., 128K tokens). Since our model is trained on \textbf{NarrativeQA}~\cite{kovcisky2018narrativeqa}, we conduct in-domain evaluation on its held-out books, and perform out-of-domain evaluation on the EN.MC subset of \textbf{$\infty$Bench}~\cite{DBLP:conf/acl/ZhangCHXCH0TW0024}, the public subset of \textbf{Nocha}~\cite{DBLP:conf/emnlp/KarpinskaTLGI24}, and \textbf{DetectiveQA-En/Zh}~\cite{xu2025detectiveqa}.

Dataset statistics, baselines, experimental details, and computational cost are provided in Appx.~\ref{sec:implenatation_details}.

\section{Experiments}

\subsection{Study I: Retrieval Results}
\label{sec:retri_result}

Table~\ref{tab:recall_result} summarizes the retrieval performance.
MiA-Emb consistently outperforms all baselines across the benchmarks
and even surpasses \textbf{SitEmb}~\cite{wu2025sitemb}, a state-of-the-art
embedding model specialized for story understanding.
MiA-Emb achieves strong gains on in-domain NarrativeQA and transfers
well to out-of-domain DetectiveQA, demonstrating its ability to leverage
global context for accurate cross-domain evidence localization.

\paragraph{Comparison with Strong Embedding Models.}
To further assess the effectiveness of mindscape-aware retrieval, we compare
MiA-Emb with diverse embedding models on the out-of-domain DetectiveQA-ZH
benchmark, including GTE-Qwen2-7B-Instruct~\cite{li2023towards},
NV-Embed-v2-7B~\cite{lee2025nvembedimprovedtechniquestraining},
Voyage-Context-3~\cite{voyage2025context3}, and
SitEmb-8B~\cite{wu2025sitemb}.
We additionally include SFT-Emb-8B, which is trained with the same
supervision as MiA-Emb but without mindscape conditioning.
As shown in Table~\ref{tab:embedding_models}, MiA-Emb achieves the best
performance across all retrieval depths, outperforming both strong
out-of-the-box models and trained baselines.
Notably, it improves average recall from 51.1 to 59.5 over SFT-Emb,
highlighting the benefit of incorporating global mindscape information
beyond supervised fine-tuning alone.

\begin{table}[t]
\centering
\scriptsize
\setlength{\tabcolsep}{5pt}
\renewcommand{\arraystretch}{1.0}
\resizebox{\columnwidth}{!}{
\begin{tabular}{lcccc}
\toprule
\textbf{Model} & \textbf{R@3} & \textbf{R@5} & \textbf{R@10} & \textbf{Avg.} \\
\midrule
\multicolumn{5}{l}{\textbf{Out-of-box}} \\
Qwen3-Embedding-8B & 28.6 & 39.1 & 55.6 & 41.1 \\
NV-Embed-v2-7B & 17.5 & 24.7 & 40.7 & 27.6 \\
GTE-Qwen2-7B-Instruct & 21.0 & 30.4 & 48.3 & 33.2 \\
Voyage-Context-3$^{\dagger}$ & 36.1 & 46.8 & 63.3 & 48.7 \\
\midrule
\multicolumn{5}{l}{\textbf{Trained}} \\
SitEmb-8B$^{\dagger}$ & 42.5 & 54.5 & 69.3 & 55.4 \\
SFT-Emb-8B & 37.9 & 48.8 & 66.5 & 51.1 \\
\rowcolor{gray!10}
\textbf{MiA-Emb-8B} &
\textbf{46.8} & \textbf{59.2} & \textbf{72.5} & \textbf{59.5} \\
\bottomrule
\end{tabular}
}
\caption{Retrieval performance on DetectiveQA-ZH.
$^{\dagger}$ denotes results reported in SitEmb~\cite{wu2025sitemb}.}
\label{tab:embedding_models}
\end{table}

\paragraph{Generalization to Long-Term Conversational Memory.}
We further evaluate MiA-Emb on LoCoMo~\cite{LoCoMo}, a long-term
conversational memory benchmark substantially different from NarrativeQA.
Without retraining, MiA-Emb consistently improves over Qwen3-Embedding-8B,
with Recall@3/5/10 increasing from
58.61/67.67/79.15 to 74.23/81.84/89.18, respectively.
These results demonstrate that the learned mindscape-aware representations
generalize beyond long-narrative understanding to long-term conversational
memory retrieval.

\subsection{Study II: Long Narrative Understanding}
\subsubsection{End-Task Results}
Table~\ref{tab:qa_overall_avg} presents the complete RAG pipeline evaluation across five long-context benchmarks. 
We also include results on the Helmet~\cite{yen2024helmet} version of NarrativeQA to compare with long-window LLM solutions in Appx.~\ref{appendix:helmet}, on which our MiA-RAG improves over GPT-4o with only $\sim$3\% of input tokens.
There are three key takeaways.


\paragraph{MiA-RAG Achieves Superior Overall Performance.}
MiA-RAG attains the best results across all benchmarks, spanning English and Chinese, diverse domains, and multiple task formats, with gains of +16.18\% over vanilla-RAG-14B and +8.63\% over vanilla-RAG-72B, showing the effectiveness of mindscape capability for RAG systems.

\paragraph{Mindscape-Aware Retrieval Consistently Improves Performance}
As verified in Sec.~\ref{sec:retri_result}, MiA-Emb substantially enhances retrieval quality. When integrated into the full pipeline, 
substituting the vanilla retriever with MiA-Emb yields consistent gains.
MiA-Emb improves the average scores of the 72B and 14B generators by 6.95\% and 7.55\%, respectively, confirming that globally informed queries yield more effective retrieval.

\paragraph{Integrative Reasoning Benefits from Mindscape-Conditioned Generation}
Simply supplying the global summary to an off-the-shelf generator improves the overall performance at both generator scales, indicating that global contextual cues are useful even without additional training.
More importantly, under the same retriever, retrieved chunks, and summary input, MiA-Gen-14B achieves a clear further improvement over its vanilla Qwen2.5-14B counterpart.
This result suggests that mindscape-conditioned fine-tuning enables the generator to use the summary as a semantic scaffold and more effectively integrate retrieved evidence within the global context.

\begin{table}[t]
\centering
\renewcommand{\arraystretch}{1.2} 
\setlength{\extrarowheight}{0.7pt} 
\resizebox{\linewidth}{!}{
\begin{tabular}{l|ccc|ccc|ccc}
\toprule
\multirow{2}{*}{\textbf{Method}}
& \multicolumn{3}{c|}{\textbf{NarrativeQA}}
& \multicolumn{3}{c|}{\textbf{DetectiveQA-ZH}}
& \multicolumn{3}{c}{\textbf{DetectiveQA-EN}} \\
\cmidrule(lr){2-4}
\cmidrule(lr){5-7}
\cmidrule(lr){8-10}
& 3 & 5 & 10
& 3 & 5 & 10
& 3 & 5 & 10 \\
\midrule

\textbf{MiA-Emb-8B}
& \textbf{62.68}
& \textbf{75.92}
& \textbf{88.09}
& \textbf{46.75}
& \textbf{59.17}
& \textbf{72.50}
& \textbf{42.08}
& \textbf{54.17}
& \textbf{69.75} \\

\quad w/o Summary
& 55.62 & 67.19 & 83.65
& 37.92 & 48.75 & 66.50
& 34.00 & 45.75 & 61.25 \\

\midrule
SitEmb-8B
& 59.98 & 70.70 & 82.68
& 42.50 & 54.50 & 69.30
& 36.75 & 49.25 & 63.83 \\

Qwen3-Embedding-8B
& 41.81 & 54.51 & 71.13
& 28.58 & 39.08 & 55.58
& 24.17 & 34.17 & 49.25 \\

\bottomrule
\end{tabular}
}
\caption{Retrieval performance measured by Recall (\%).}
\label{tab:recall_result}
\end{table}

\begin{table}[t]
\centering
\scriptsize
\renewcommand{\arraystretch}{0.98}
\begin{tabular}{lcccc}
\toprule
\textbf{Method} & \textbf{Narra.QA} & \textbf{$\infty$Bench} & \textbf{DetectiveQA} & \textbf{NoCha} \\
\midrule
MiA-Gen-14B& \textbf{53.19} & \text{82.39} & \textbf{76.03} & \textbf{52.91} \\
\hline
w/o Summary & 50.49  & 75.39  & 71.03 & 44.97  \\
w/o Claim. & 51.22 & \textbf{85.44} & 75.58 & 44.44 \\
w/o QA & 46.40 & 81.08 & 72.81 & 46.56 \\
\bottomrule
\end{tabular}
\caption{Ablation study of \text{MiA-Gen-14B}. 
Reported scores are averaged over the 3/5/10-chunk settings.}
\label{tab:gen_ablation}
\end{table}

\subsubsection{Ablation Study}

We perform ablations to assess each MiA-RAG component, with results shown in Tables~\ref{tab:recall_result} and \ref{tab:gen_ablation}.

\noindent\textbf{Impact of Mindscape-Conditioning }
Same as in the embedding stage (Table~\ref{tab:recall_result}), removing the summary (\textit{w/o Summary}) leads to substantial degradation in the generation stage (Table~\ref{tab:gen_ablation}). 
These declines indicate that summary-based supervision helps align queries with global semantics and supports the integration of dispersed evidence.

\noindent\textbf{Benefit of Multi-Paradigm Supervision }
Ablating either supervision paradigm (\textit{w/o Claim.} for claim verification or \textit{w/o QA} for question answering) consistently degrades performance, indicating that exposure to diverse reasoning patterns improves generalization beyond any single task.
\begin{table}[t]
\centering
\resizebox{0.95\linewidth}{!}{
\begin{tabular}{lcc}
\toprule
\textbf{Summary Generator} & \textbf{Recall@3/5/10 (\%)} & \textbf{F1-Score (\%)} \\
\midrule
GPT-4o (Ours) & 62.68/75.92/88.09 & 52.48/53.52/53.56 \\
\hline

Qwen2.5-32B-Instruct &61.66/74.60/88.06 & 50.20/51.80/53.37\\
Qwen2.5-14B-Instruct & 59.74/73.54/87.68 & 51.45/51.81/52.61 \\
Qwen2.5-7B-Instruct & 58.62/72.61/86.17 & 49.79/51.55/51.48 \\

\bottomrule
\end{tabular}}
\caption{Impact of summary quality on NarrativeQA.}
\label{tab:summary_quality}
\end{table}


\subsection{Study III: MiA-GraphRAG for Global QA}
\label{appendix:sensemaking}
\paragraph{Global Sense-Making QA Task}Beyond local evidence–oriented evaluation, we assess global sense-making questions that require a holistic understanding of the entire document. These questions are constructed from the LongBench~\cite{bai2023longbench} summary-generation datasets: QMSum and GOV (English), and VCSum (Chinese). Each question is derived from source documents exceeding 100K tokens, ensuring that the model must integrate global information rather than rely on localized evidence. In total, we construct 300 such questions. Prompt is provided in Figure~\ref{fig:prompt_sensemaking_question_generation}.

\paragraph{Results}
We evaluate global sense-making in a GraphRAG QA setting. Three node retrievers are compared for selecting semantic entities from the document-level knowledge graph: (1) our \texttt{MiA-Emb}, (2) a multi-task embedding model trained without mindscape supervision (\texttt{SFT-Emb}), and (3) the \texttt{vanilla} Qwen3-Embedding-8B. Each retriever selects the top-20 nodes, after which their associated relations and supporting chunks are assembled into the global semantic context following the local mode of GraphRAG procedure~\cite{edge2024local}.

We conduct pairwise comparisons judged by GPT-4o along three dimensions: \textbf{Comprehensiveness}, \textbf{Diversity}, and \textbf{Empowerment} (Figure~\ref{prompt:sense_making_eval} in Appx~\ref{appendix:prompt}). As shown in Table~\ref{tab:sensemaking_results}, MiA-Emb achieves the best performance across all dimensions under the same graph construction pipeline. This indicates that mindscape-aware retrieval surfaces entities that more accurately capture the document’s overall semantic structure.
\begin{table}[htb]
\centering
\scriptsize
\setlength{\tabcolsep}{3pt}
\renewcommand{\arraystretch}{0.95}

\begin{tabular}{
    l|ccc|ccc
}
\toprule
\multirow{2}{*}{\textbf{Dimension}}
& \multicolumn{3}{c|}{\textbf{(A) \texttt{MiA-Emb} vs \texttt{SFT-Emb}}}
& \multicolumn{3}{c}{\textbf{(B) \texttt{MiA-Emb} vs \texttt{Vanilla}}} \\
\cmidrule(lr){2-4}\cmidrule(lr){5-7}
& \textbf{A1} & \textbf{A2} & \textbf{Win}
& \textbf{A1} & \textbf{A2} & \textbf{Win} \\
\midrule
Comprehensiveness
& 87.74 & 12.26  &\texttt{MiA-Emb}
& 88.39 & 11.61 & \texttt{MiA-Emb} \\

Diversity
& 68.39 & 31.61 & \texttt{MiA-Emb}
& 63.23 & 36.77 & \texttt{MiA-Emb} \\
Empowerment
& 73.87 & 26.13 & \texttt{MiA-Emb}
& 71.94 & 28.06 & \texttt{MiA-Emb} \\
Overall Winner
& 81.29 & 18.71 & \texttt{MiA-Emb}
& 78.39 & 21.61 & \texttt{MiA-Emb} \\
\bottomrule
\end{tabular}

\caption{Pairwise comparison of MiA-based methods vs baselines across evaluation dimensions. Values are percentages. We use Qwen2.5-72B as the generator. }
\label{tab:sensemaking_results}
\end{table}

\subsection{Study IV: Impact of Model Scales}
\label{sec:scale_model}
\begin{figure}[t]
    \centering
    \includegraphics[width=0.99\linewidth]{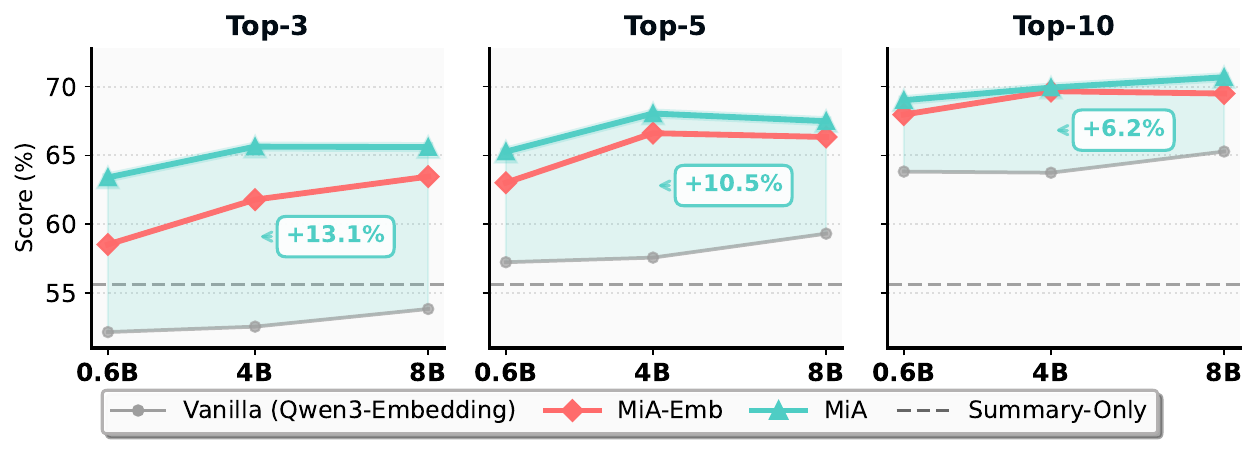}
     \caption{Impact of retriever scale on average results across benchmarks (DetectiveQA-ZH/EN,$\infty$Bench, NoCha, NarrativeQA) with a Qwen2.5-72B generator.}
    \label{fig:scaled_emb_results}
\end{figure}

\begin{figure}[t]
    \centering
    \includegraphics[width=0.995\linewidth]{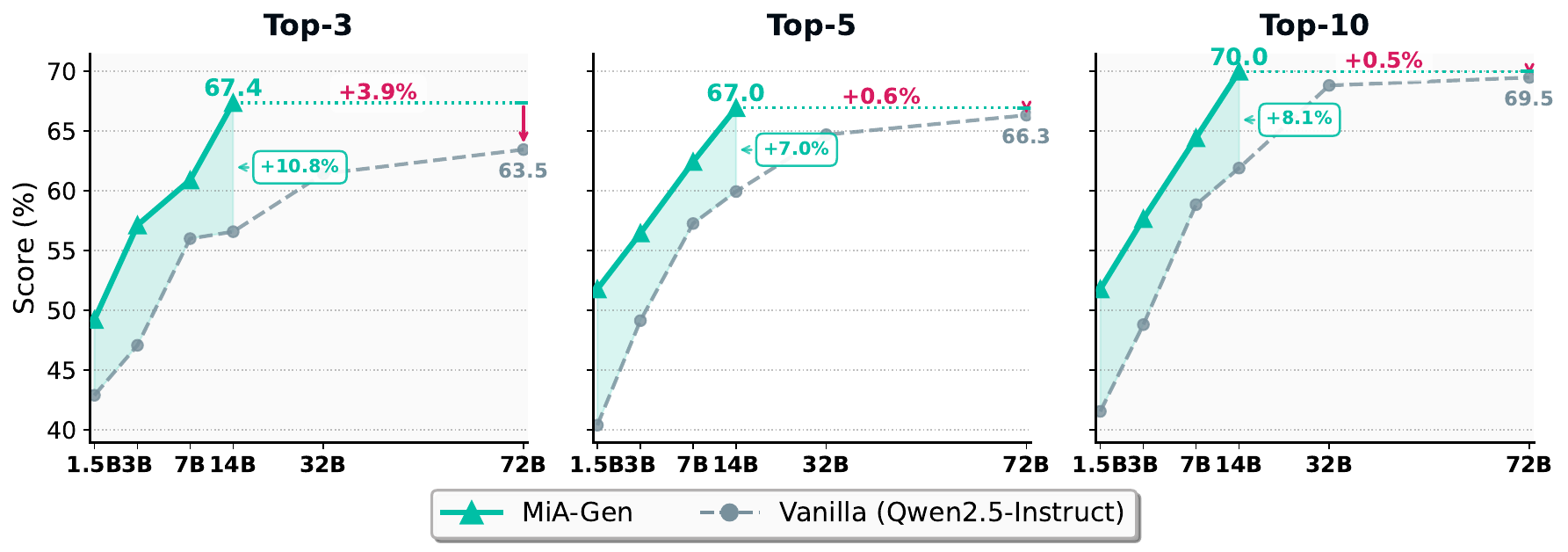}
    \caption{Impact of generator model scale on average results over 5 benchmarks,  with a MiA-Emb-8B retriever.}
    \label{fig:scaled_gen_results}
\end{figure}

We evaluate the scalability of MiA-Emb across backbone sizes (0.6B to 8B) against \texttt{SFT-Emb} (identical to the \textit{w/o Summary} ablation in Sec.~\ref{sec:retri_result}, i.e., trained and evaluated without summaries) and \texttt{Vanilla} Qwen3-Embedding baselines.
As shown in Figures~\ref{fig:scaled_emb_results} and \ref{fig:scaled_recall_results}, MiA-Emb consistently outperforms both baselines; notably, \text{MiA-Emb-0.6B} already surpasses the \text{Vanilla 8B} model in both retrieval recall and downstream QA and reasoning performance. 
We also scale MiA-Gen across model sizes and observe consistent gains over the vanilla Qwen2.5-Instruct models (1.5B$\sim$72{B}), presented in Figure~\ref{fig:scaled_gen_results}. In particular, \textbf{MiA-Gen-14B}  even surpasses the {72B} model. These results demonstrate that our approach exhibits scalability across model sizes.
Numerical results are provided in Appx.~\ref{appendix:scale_model}.

\subsection{Study V: Impact of Summary Quality}
We examine the robustness of MiA-RAG to summary quality by replacing GPT-4o summaries with those generated by open-source Qwen2.5 models (7B, 14B, and 32B).
As shown in Table~\ref{tab:summary_quality}, MiA-RAG maintains stable performance across summarizers, with the 14B model achieving results comparable to GPT-4o.
It is also robust to summary length: NarrativeQA summaries range from 240 to 1,243 tokens but show only a weak correlation with Recall@10 (Pearson $r=0.23$).
These results indicate that summary information loss has a limited impact, since the mindscape serves as a global scaffold for retrieval and reasoning rather than a lossless store of fine-grained evidence.
We further validate this observation in Sec.~\ref{ssec:analysis_summ_use} and Sec.~\ref{ssec:geometric_properties}.
\section{Analysis}
\label{sec:analysis}
In this section, we introduce analytical methods to evaluate whether the resulting MiA-RAG exhibits the three hypothesized capabilities proposed in the introduction, namely \emph{Enriched Understanding}, \emph{Selective Retrieval}, and \emph{Integrative Reasoning}.

\subsection{The Role of Global Summaries}
\label{ssec:analysis_summ_use}
While ablation confirms that MiA-RAG benefits from incorporating summaries, a key question remains: what is their functional role during inference?
We first show that summaries are not useful simply because they cover the answer. We evaluate a \emph{Summary-Only} variant where the generator predicts answers using only the summary (Table~\ref{tab:qa_overall_avg}). This variant consistently underperforms vanilla-RAG and falls far short of MiA-RAG.
These results, together with same-summary retrieval controls (Table~\ref{tab:vanilla_model_residual}), indicate that summaries function not as standalone evidence or auxiliary text, but as \emph{semantic scaffolds} for retrieval and reasoning.

\subsection{Geometric Properties of the MiA Embedding Space}
\label{ssec:geometric_properties}
Extending the Sec.~\ref{ssec:analysis_summ_use}, we further examine:

\rqsection{MiA-Emb  facilitates \text{Selective Retrieval}.}

That is, whether the embedding model biases query representations toward the active book topic, thereby better positioning them within the subspace supported by the corresponding chunks.

\paragraph{Method}
We visualize query and chunk embeddings with t-SNE~\cite{maaten2008visualizing}. 
To characterize the semantic structure of the document,
We first fit t-SNE on the chunk embeddings only, yielding a 2D manifold that reflects the document’s semantic structure. 
We then embed the query representations into the same 2D space and inspect how well each model positions queries relative to the corresponding topic-relevant chunk regions.

\begin{figure}
    \centering
    \includegraphics[width=0.99\linewidth]{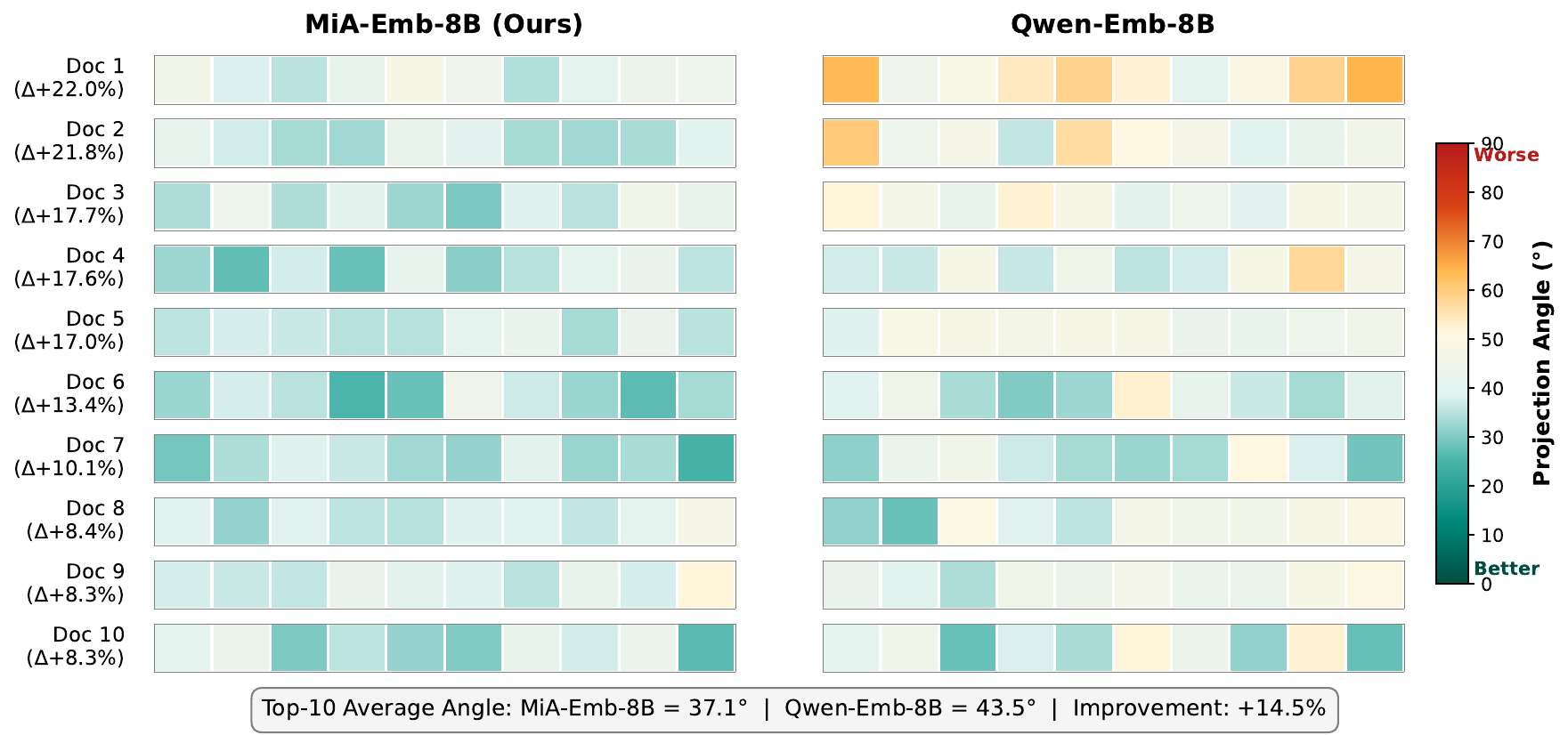}
    \caption{Comparison of projection angles for MiA-Emb and Qwen3-Emb. Lower angles indicate better alignment of queries with the book's semantic subspace.}

\label{fig:projection_angles}
\end{figure}

\paragraph{Results}
Figure~\ref{fig:projection_angles} shows a clear geometric distinction between MiA-Emb and the vanilla embedding model. Across books, MiA-Emb consistently yields smaller projection angles, meaning that query embeddings lie closer to the semantic subspaces spanned by their corresponding documents. 
On average, MiA-Emb-8B achieves 37.1°, compared with 43.5° for Qwen-Emb-8B, demonstrating that mindscape conditioning more effectively guides queries toward the correct semantic region and enables more precise selective retrieval.

\subsection{Residual Stream and Attention Analysis of the MiA Embedding Model}

We analyze the model's internal representations and attention  to examine the following hypothesis:

\rqsection{MiA-Emb facilitates \textbf{Enriched Understanding} of queries.}

We verify the hypothesis in two folds. First, we examine whether performance gains from MiA-Emb correlate with increased use of the global summary. If so, we then study whether the model focuses its attention on information that can enrich the queries in these situations.
\begin{figure}[!htb]
    \centering
    \includegraphics[width=0.99\linewidth]{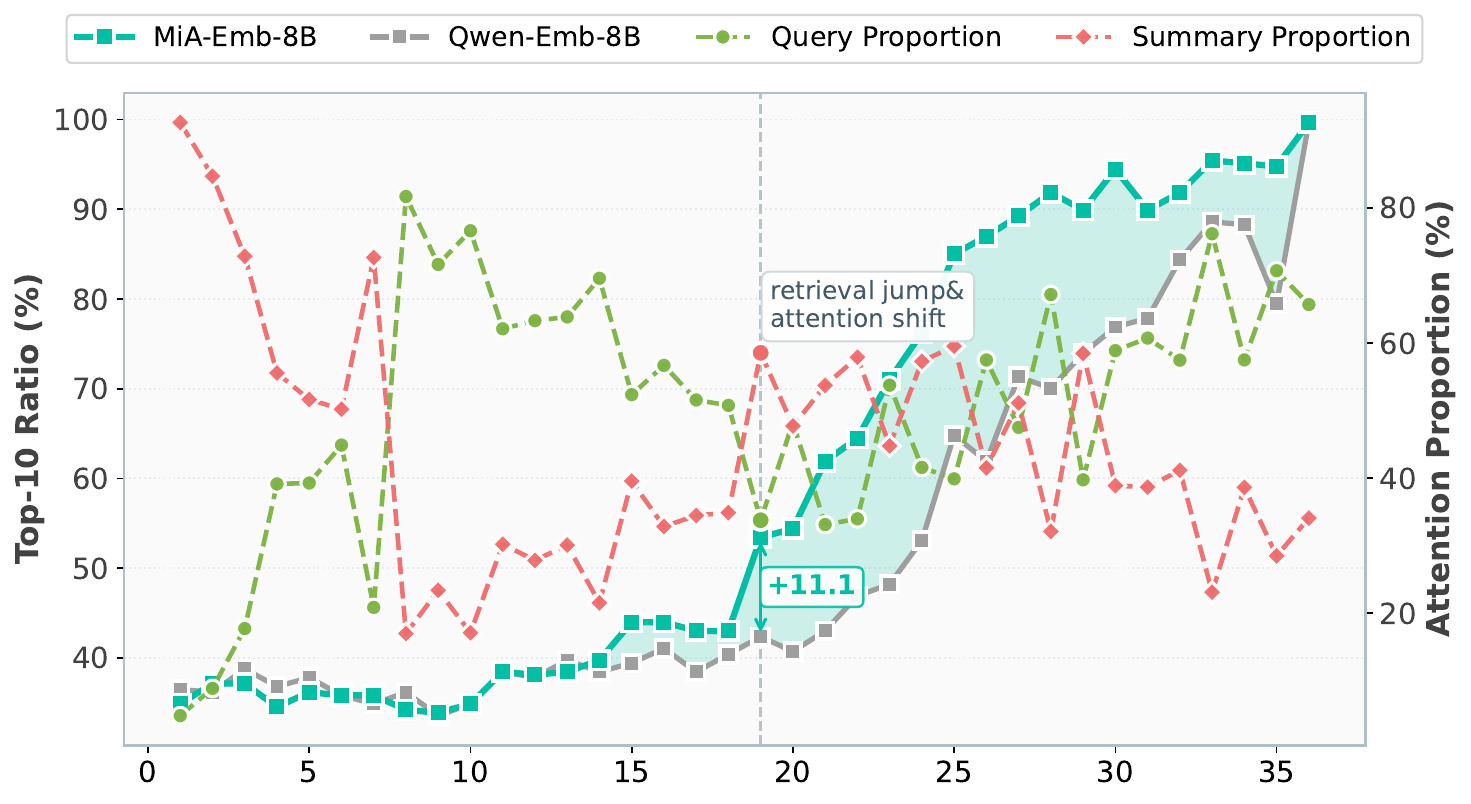}
    \caption{Layer-wise comparison of silver-chunk retrieval accuracy and attention allocation proportion.}
    \label{fig:emb_residual}
\end{figure}

\rqsous{MiA-Emb puts increased attention to the global summary at layers of improved predictability compared to the baseline.}
\label{rq:embedding_attention}

\paragraph{Method}
Following the approach of~\cite{jiang2024large}, we compare MiA-Emb with the vanilla embedding model through their residual streams to analyze how retrieval-relevant information is progressively accumulated into the query representation. To ensure comparability, we select 100 queries for which both models achieve Recall@10 = 100\%.
Concretely, we track the layer-wise Top-10 silver-chunk ratio for both models, reflecting how the hidden states at each layer steer the retrieval distribution toward the correct evidence.
For MiA-Emb, we additionally examine the attention from the last token to summary tokens and to query tokens, enabling us to assess whether improvements coincide with increased use of global-summary cues.

\paragraph{Results} 
As shown in Figure~\ref{fig:emb_residual}, MiA-Emb exhibits a clear rise in silver-chunk recall beginning at the middle layers. This rise coincides with increased attention to the global summary in the same layer range, suggesting that the model progressively injects summary-derived cues into the query representation. This incorporation of global signals enriches the query embedding, enabling MiA-Emb to develop a deeper semantic understanding of the query and thus support more selective retrieval.

\rqsous{MiA-Emb attends to information that enriches the query at the layers identified in \ref{rq:embedding_attention}.}
\paragraph{Method}
To understand how summary information enriches query representations, we inspect the summary-attentive layers identified in Section~\ref{rq:embedding_attention}.
Our goal is to assess whether the embedding token allocates its attention to summary tokens that are semantically aligned with the query.
If such attention emerges precisely at layers where retrieval performance improves, it suggests that MiA-Emb enhances query understanding through targeted integration of global context.
\begin{figure}[t]
    \centering
    \includegraphics[width=0.95\linewidth]{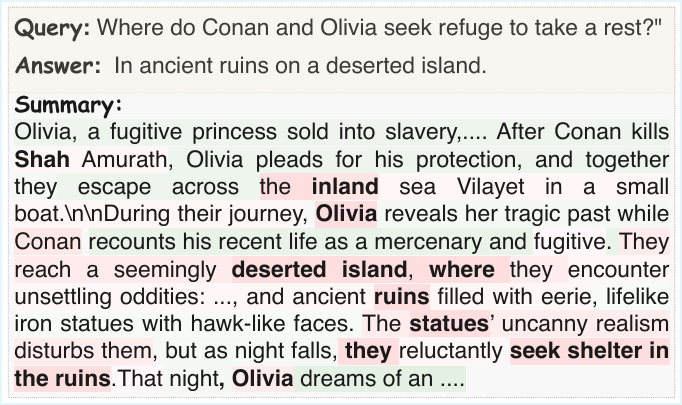}
    \caption{Attention pattern of MiA-Emb: the last token attends to preceding summary tokens, with red regions indicating tokens that receive high attention.}
    \label{fig:emb_attn_words}
\end{figure}

\paragraph{Results}
Figure~\ref{fig:emb_attn_words} shows that, at the layers corresponding to retrieval gains, the final embedding token concentrates its attention on summary phrases that correspond to answer-relevant entities, events, or locations.
This indicates that MiA-Emb selectively extracts semantically aligned global cues and integrates them into the query representation, reinforcing the layer-wise analysis and confirming that MiA-Emb enhances query understanding via summary-based enrichment.

\subsection{Attention Pattern Analysis in the Generation Model}
\label{ssec:attention_pattern}

\begin{figure}[t]
    \centering
    \includegraphics[width=1\linewidth]{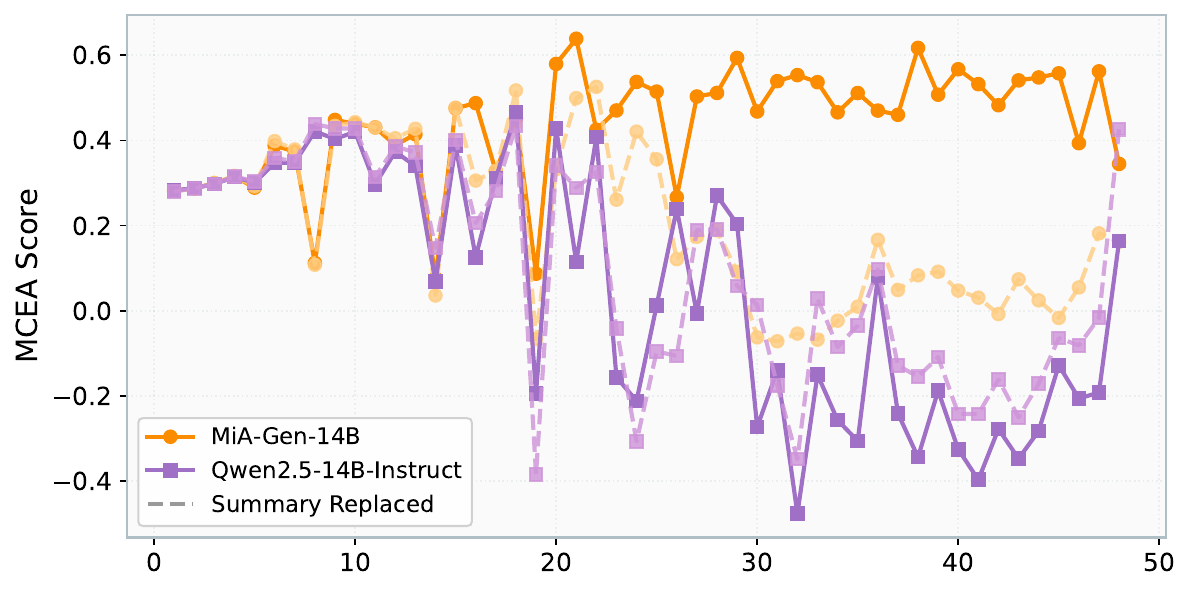}
    \caption{Layer-wise Mindscape-Coherent Evidence Alignment (MCEA) scores for generator. }
    \label{fig:mia_gen_mcea}
\end{figure}

\rqsection{MiA-Gen facilitates \textbf{Integrative Reasoning} over retrieved chunks within the global mindscape.}

To examine how the mindscape steers generation toward relevant evidence, we introduce the Mindscape-Coherent Evidence Alignment metric (MCEA).
MCEA measures whether chunks receiving stronger mindscape attention are also preferentially attended by the query, comparing this coupling between relevant and noise chunks.
Thus, MCEA serves as an attention-level diagnostic of mindscape-conditioned evidence integration, rather than a raw score for direct cross-model comparison.
A formal definition is given in Appx.~\ref{appendix:mcea_definition}.
\paragraph{Method}
We compute MCEA for MiA-Gen and Qwen2.5-14B-Instruct at each transformer layer.
To test whether the observed coupling depends on valid global information rather than incidental attention, or positional bias, we introduce a \emph{summary-replaced} control that replaces the original summary with an unrelated text of the same length.


\paragraph{Results}
Figure~\ref{fig:mia_gen_mcea} highlights two key findings.
First, MiA-Gen exhibits consistently high MCEA values, especially in middle and late layers.
This suggests that local chunks first absorb global mindscape information, after which the query increasingly attends to the enriched evidence.
Second, replacing the valid summary with unrelated text sharply reduces MiA-Gen's MCEA, showing that its evidence integration depends on valid mindscape.
In contrast, the vanilla model shows little sensitivity to this perturbation; its raw MCEA under unrelated summaries can remain comparable, likely reflecting incidental attention coupling rather than meaningful mindscape-to-evidence transfer.

\section{Conclusion}
Inspired by human cognitive ability to interpret inputs within a global mindscape, we propose MiA-RAG,  a mindscape-aware framework for LLM-based RAG.
MiA-RAG approximates this global impression with hierarchical summaries as persistent global memory, conditioning both retrieval and generation on it.
MiA-RAG achieves strong performance on evidence-based long-context understanding and global sense-making tasks.
Empirical analysis shows that the mindscape supports global semantic query understanding, selective retrieval, and integrative reasoning over dispersed evidence.

\section*{Limitations}

While MiA-RAG demonstrates strong performance on narrative long-context QA and reasoning, it relies on a precomputed global summary as the mindscape representation, which introduces additional preprocessing overhead compared with standard RAG. Although this cost is incurred only once per document and can be amortized across subsequent queries, it remains a practical consideration for deployment. At the same time, the hierarchical bottom-up design supports incremental updates by regenerating only the affected paths, which partially mitigates this limitation in dynamic or evolving document settings. In addition, our experiments mainly focus on narrative long-context understanding, while other application settings remain unexplored. Nevertheless, the strong performance on a global sense-making dataset provides initial evidence that MiA-RAG may generalize beyond purely narrative QA settings. Finally, part of the supervision signal is derived from commercial LLMs, which may introduce latent bias or hallucinated content. Nevertheless, the empirical results suggest that the mindscape-aware training strategy remains robust under imperfect supervision.

\section*{Acknowledgments}
This work was supported by the National Natural Science Foundation of China (Nos. 62472419,
62472420) and the Enterprise Project (No.
E4V06811F3).
\bibliography{custom}

\appendix

\appendix

\section{Supports of Mindscape-Aware Capabilities in Broader Research Fields}
\label{sec:supports}
We show that the existence and advantages of mindscape-aware capabilities are supported by research on human memory in psychology and neuroscience.

\paragraph{Supports in Psychology}
The existence of mindscape-aware capability traces back to the concept of schema~\cite{bartlett1932remembering} and aligns with Fuzzy-Trace Theory (FTT; \citealt{reyna1995fuzzy}).
Schemas provide integrated structures for familiar topics, guiding attention and constraining interpretation during information processing.
FTT further posits that human memory encodes experiences at two complementary levels: verbatim traces that preserve surface details and gist traces that capture abstract, meaning-based structure.
When new information relates to a familiar topic, gist-level representations are reactivated, providing a global semantic scaffold for interpretation, retrieval, and reasoning.
Our mindscape-aware framework can be viewed as a computational approximation of such gist-based cognition in retrieval-augmented reasoning systems.

The advantages of mindscape-aware capability are also related to the Encoding Specificity Principle~\cite{tulving1973encoding}.
This principle states that reinstating the original contextual pattern of a familiar topic or task can reactivate the corresponding memory network, thereby improving retrieval effectiveness and interpretive coherence.

\paragraph{Supports in Neuroscience}
Mindscape-aware capabilities are also supported by neuroscience research.
The \emph{controlled semantic cognition (CSC)} framework~\cite{ralph2017neural} suggests that cognition is guided by globally integrated semantic knowledge, allowing thought to unfold within a coherent knowledge framework.

Neuroimaging studies provide further evidence for these mechanisms.
Prior work identifies neural foundations of schemas that support new knowledge integration and shape memory recall~\cite{brod2017neural,gilboa2017neurobiology,audrain2022schemas}.
Other findings show that context reinstatement during retrieval, \emph{i.e.}, reactivating semantic, situational, or cue-related features present during encoding, enhances memory recall~\cite{gershman2013neural,kragel2021distinct}.
Similarly, the CSC framework is supported by evidence of sustained co-activation during story comprehension and semantic processing~\cite{binder2009semantic,xu2016intrinsic}.
In this sense, mindscape-awareness can be viewed as a higher-order manifestation of encoding specificity and schema mechanisms, where a global semantic mindscape guides interpretation and retrieval.

\section{MiA-Emb: Supervision and Training Objective}
\label{appendix:silver_annotate}
\begin{algorithm}[t]
\caption{Silver Evidence Annotation}
\label{alg:silver_evidence}
{\small
\begin{algorithmic}[1]
\State \textbf{Input:} Dataset $\mathcal{D}=\{(q_i,a_i)\}_{i=1}^N$, mindscape summary $S$, retriever $\mathcal{E}_s$, task $t\in\{c,n\}$.
\State \textbf{Define:} Evidence units $U = C$ (if $t=c$) or $U = V$ (if $t=n$).
\State \textbf{Output:} Silver-annotated dataset $\tilde{\mathcal{D}}^t_{\text{emb}}=\{(q_i, \tilde{U}_i)\}_{i=1}^N$
\State Initialize $\tilde{\mathcal{D}}^t_{\text{emb}} \gets \emptyset$
\For{each $(q_i,a_i)$ in $\mathcal{D}$}
    \State \Comment{\textit{Query Augmentation}}
    
    \State $q_{\text{aug}} \gets \{q_i,\; q_i+a_i, \;a_i\},$
    \State \Comment{\textit{Candidate Retrieval \& Ensemble}}
    \State $U_{\text{pool}} \gets \emptyset$
    \For{$q'$ in $q_{\text{aug}}$}
        \State $U_{\text{pool}} \gets U_{\text{pool}} \cup \text{RetrieveTopK}(\mathcal{E}_s, q', U, k)$
    \EndFor
    \State $U_{\text{cand}} \gets \text{VoteAndSelectTopK}(U_{\text{pool}}, k)$
    \State \Comment{\textit{LLM-driven Refinement}}
    \State $\tilde{U}_i \gets \text{LLM}_t(q_i, a_i, U_{\text{cand}})$ \Comment{See Fig.~\ref{prompt:filter_silver_chunks}/\ref{prompt:filter_silver_nodes}}
    \State Add $(q_i, \tilde{U}_i)$ to $\tilde{\mathcal{D}}^t_{\text{emb}}$
\EndFor
\State \Return $\tilde{\mathcal{D}}^t_{\text{emb}}$
\end{algorithmic}
}
\end{algorithm}
\subsection{Positive Evidence  Construction} 
\label{appendix:positive_evidence}
As existing long-context benchmarks lack explicit query–evidence alignments, we automatically construct silver evidence for both chunk- and node-level retrieval.

\label{appendix:silver_node}
\paragraph{Silver Chunk Annotation}
We annotate \emph{silver chunks} using a structured procedure that integrates query augmentation, ensemble retrieval, and LLM-based refinement (Algorithm~\ref{alg:silver_evidence}). For chunk-level supervision, we set the task $t = c$ in Algorithm~\ref{alg:silver_evidence}, yielding the silver chunk dataset $\tilde{\mathcal{D}}^c_{\text{emb}} = \{(q_i, \tilde{C}_i)\}_{i=1}^N$, where $\tilde{C}_i \subset C$ denotes the set of supporting chunks for query $q_i$.
\paragraph{Silver Node Annotation} 
\label{appendix:silver_node}
To support retrieval at a global semantic granularity, we construct a knowledge graph $G=(V, E)$ by extracting entity-level information, following a procedure similar to GraphRAG~\cite{edge2024local}. 
For each document, we employ GPT-4o to identify key entities and generate concise descriptions, yielding a node set $V = \{(e^{\text{name}} : e^{\text{desc}})\}$.

We then generate the node-level silver dataset $\tilde{\mathcal{D}}^n_{\text{emb}} = \{(q_i, \tilde{V}_i)\}_{i=1}^N$ by setting the task $t=n$ and evidence units $U=V$ in Algorithm~\ref{alg:silver_evidence}. 
Here, $\tilde{V}_i \subset V$ represents the set of relevant nodes for query $q_i$, serving as the ground truth for the node retrieval task.

\subsection{Negative Evidence Construction}
\label{appendix:negative}
MiA-Emb is trained with a contrastive objective that requires both positive and negative samples. Positive samples are taken from the silver evidence sets $\tilde{C}_i$ and $\tilde{V}_i$ described above, while negative samples are constructed from two complementary sources. We illustrate the construction for chunk retrieval; node retrieval follows the same design.
\noindent\textbf{Hard negatives.}  
Hard negatives are semantically similar to the query but not included in the silver evidence. We select chunks from the candidate set $C_{\text{cand}}$ (Algorithm~\ref{alg:silver_evidence}) that are not part of the silver set $\tilde{C}_i$, and take up to $5$ such chunks to form the hard-negative set $C^{\text{hard}}_i$. These samples provide challenging contrasts that encourage the model to distinguish subtle semantic differences.

\noindent\textbf{Simple negatives.}  
Simple negatives are clearly irrelevant to the query. We sample them uniformly at random from the full document chunk set $C$, ensuring no overlap with the positive set $\tilde{C}_i$ or the hard negatives $C^{\text{hard}}_i$. We sample $5$ chunks to form the simple-negative set $C^{\text{simple}}_i$.

\noindent\textbf{Final negative set.}  
For chunk retrieval, the final negative pool for query $q_i$ is
\begin{equation}
\small
C^{-}_i = C^{\text{hard}}_i \cup C^{\text{simple}}_i.
\end{equation}

For node retrieval, we apply the same procedure to obtain the node-level negative set $V^{-}_i$.  we use $U^{-}_i$ as a unified notation for the negative set of query $q_i$.
\subsection{Model Training}
\label{appendix:mia_emb_opt}
We provide additional training details for MiA-Emb, complementing Sec.~\ref{sec:mia_emb}.

\paragraph{Input Representation}
To enable the embedding model to perceive both the local query intent and the global mindscape, we construct a composite input sequence. Let $q_i$ be the query and $S$ be the mindscape summary. The input is formatted as
\begin{equation}
    Q = [\textcolor{gray}{\texttt{[INST]}_{\texttt{emb}}};\ q_i;\ \textcolor{gray}{d_q};\ S;\ \textcolor{gray}{d_t}],
\end{equation}
where $\textcolor{gray}{\texttt{[INST]}_{\texttt{emb}}}$ is the instruction prefix, $\textcolor{gray}{d_q}$ marks the end of the query, and $\textcolor{gray}{d_t= [d_n, d_c]}$ serve as special tokens representing node- and chunk-retrieval tasks, respectively. 

The sequence is encoded by the embedding model $\mathcal{E}$ to obtain token-level hidden states:
\begin{equation}
\small
    \mathbf{H} = \mathcal{E}(Q) = (\mathbf{h}_1, \ldots, \mathbf{h}_{|Q|}),
\end{equation}
where $\mathbf{H}$ denotes the last-layer hidden states for all tokens in $Q$.

\paragraph{Residual Integration}
To preserve the original query semantics while injecting global context, we employ a residual connection strategy. We extract the hidden state at the query delimiter ($\mathbf{h}_q$, corresponding to token $d_q$) and the hidden state at the task delimiter ($\mathbf{h}_t$, corresponding to $d_c$ or $d_n$, depending on the active task $t$). The final enriched query representation $\tilde{\mathbf{q}}_t$ is computed as
\begin{equation}
\small
\tilde{\mathbf{q}}_t = \delta \cdot \mathbf{h}_q + (1 - \delta) \cdot \mathbf{h}_t,
\end{equation}
where $\delta$ is a hyperparameter controlling the balance between local query focus and global context awareness. A detailed ablation on the role of this residual connection is provided in \ref{app:residual}.

\paragraph{Joint Contrastive Optimization}
Finally, we optimize a multi-task contrastive objective~\cite{Oord2018RepresentationLW} over chunk and node retrieval:
\begin{equation}
\small
\mathcal{L}_{\text{MiA-Emb}} = \beta \cdot \mathcal{L}_c + (1 - \beta) \cdot \mathcal{L}_n,
\end{equation}
where $\mathcal{L}_c$ and $\mathcal{L}_n$ represent the losses for chunk and node retrieval, respectively, and $\beta \in [0,1]$ balances their contribution.

Both tasks employ the InfoNCE loss. Specifically, the objective $\mathcal{L}_t$ ($t \in \{c, n\}$) is defined as:
\begin{equation}
\small
\begin{aligned}
\mathcal{L}_t = -\frac{1}{|B|} \sum_{j=1}^{|B|}
\log \frac{
    \exp\big(\text{sim}(\tilde{\mathbf{q}}_t^{\,j}, \mathbf{d}_t^{+j}) / \tau\big)
}{
    \sum_{\mathbf{d} \in \mathcal{C}_j}
    \exp\big(\text{sim}(\tilde{\mathbf{q}}_t^{\,j}, \mathbf{d}) / \tau\big)
},
\end{aligned}
\end{equation}
where $|B|$ is the batch size, $\tau$ is the temperature parameter, and $\text{sim}(\cdot,\cdot)$ denotes cosine similarity. 

The candidate set for the $j$-th query is constructed as:
\begin{equation}
\small
\mathcal{C}_j = \{\mathbf{d}_t^{+j}\} \cup U^-_j,
\end{equation}
where $\mathbf{d}_t^{+j}$ is the positive embedding sampled from the silver evidence set $\tilde{U}_j$, and $U^-_j$ is the corresponding set of negative candidates.
\section{Experimental Settings}

\label{sec:implenatation_details}
\subsection{Dataset Statistics}
Table~\ref{tab:dataset_stats} summarizes the public long-narrative understanding datasets, including the number of queries, average tokens, and evaluation metrics. We also study a more global sense-making setting and construct additional datasets based on LongBench~\cite{bai2023longbench}; see \ref{appendix:sensemaking} for details.

\begin{table}[t]
\centering
\scriptsize
\resizebox{1.\columnwidth}{!}{
\begin{tabular}{lccc}
\toprule
\textbf{Dataset} & \textbf{Queries} & \textbf{Avg. Tokens} & \textbf{Metrics} \\ 
\midrule
NarrativeQA & 556 &83k  & F1, EM, Recall\\
$\infty$Bench-EN.MC  & 229 & 184k & Accuracy \\ 
DetectiveQA  & 1,200 &118k & Accuracy, Recall \\ 
Nocha        & 126 &139k & Pairwise Acc. \\ 
\bottomrule
\end{tabular}
}
\caption{Summary of the evaluation datasets.}
\label{tab:dataset_stats}
\end{table}

\subsection{Baselines}
\label{appendix:baselines}
We compare our approach with several baseline models, categorized into {vanilla RAG}, {graph-based RAG}, and {memory-enhanced RAG}.

\paragraph{Vanilla RAG.}
We adopt an out-of-the-box dense retrieval pipeline using Qwen3-Embedding-8B to retrieve top-$k$ contexts from the corpus.
The retrieved passages are concatenated with the user query and fed into Qwen2.5-Instruct generators (14B or 72B) for answer generation.
We use the same prompt template and retrieval budget across methods for a fair comparison.

\paragraph{Graph-based RAG.}
    We compare against HippoRAG-v2~\cite{guti'errez2025}, a graph-based RAG framework that leverages structured memory and graph-based retrieval/ranking to aggregate evidence for generation.
We retrieve the top-$k$ chunks ($ k\in \{3,5,10\}$) using the same chunk size and chunking strategy as our method across all benchmarks.

\paragraph{Memory-enhanced RAG.}
MemoRAG~\cite{qian2025memorag} adopts a memory-enhanced RAG paradigm that augments retrieval with a long-term/global memory component.
It first uses a memory model to memorize the full context and build a global memory representation.
Given a query, the memory model then generates query-specific answer clues and potential draft answers to guide multi-pass retrieval.
During clue preparation and retrieval, we use the officially released \texttt{memorag-mistral-7b-inst}\footnote{\url{https://huggingface.co/TommyChien/memorag-mistral-7b-inst}} as the memory model, and BGE-M3~\cite{bge-m3} as the retriever.

\subsection{Implementation Details}

We set the chunk size to 1200 with an overlap of 100 tokens for NarrativeQA, DetectiveQA, and $\infty$Bench, and to 200 for NoCha, following the typical context length distributions of these datasets.
We build our retriever MiA-Emb by applying LoRA~\cite{hu2022lora} on top of Qwen3-Embedding-8B, and build our generator MiA-Gen by fully fine-tuning Qwen2.5-14B-Instruct.
Throughout the paper, we denote Qwen2.5-72B as the 4-bit quantized variant of Qwen2.5-72B-Instruct to improve efficiency. \textit{GPT-4o} refers to GPT-4o-2411. For silver evidence construction, we use GTE-Qwen2-7B as the retriever $\mathcal{E}_s$   and $K=10$ for top-K in Algorithm~\ref{alg:silver_evidence}. All hyperparameters are summarized in Table~\ref{tab:training_config}.
\begin{table}[t]
\centering
\small
\setlength{\tabcolsep}{6pt}
\renewcommand{\arraystretch}{1.15}
\begin{tabular}{lcc}
\toprule
\textbf{Setting} & \textbf{MiA-Emb } & \textbf{MiA-Gen} \\
\midrule

Precision          
& bfloat16 
& bfloat16 \\
Batch Size         
&  4 
& 2 \\
Steps     
& 2000 
& 2000 \\
warmup ratio
&0.1  &0.05 \\
Learning Rate      
& $1\times10^{-4}$  
& $1\times10^{-5}$ \\
LoRA Rank          
& 128 
& – \\
LoRA $\alpha$      
& 256 
& – \\
Temperature $\tau$     
& 0.01 
& 0 \\
Residual Weight $\delta$ 
& 0.5 
& – \\

\bottomrule
\end{tabular}
\caption{Training configurations.`-' denotes not used.}
\label{tab:training_config}
\end{table}
\subsection{Computational Cost}
\label{appendix:cost}

\paragraph{Training Cost}
MiA-Emb (8B) is trained for approximately 21 hours on 8$\times$H20 GPUs ($\approx$168 GPU-hours). MiA-Gen (14B) is trained for approximately 28 hours on the same hardware ($\approx$224 GPU-hours).

\paragraph{Mindscape Construction Cost}
Using vLLM with Qwen2.5-7B-Instruct on 2$\times$A100(40GB) GPUs, the average mindscape build time is approximately 24.79 seconds per book (average 118k tokens on DetectiveQA-En). This is a one-time preprocessing cost amortized across all downstream queries for the same document.

\paragraph{End-to-End Inference Latency}
Concatenating the summary increases retrieval latency (from 10.48 to 100.23 ms/query on $\infty$Bench) but improves evidence selection, reducing the generation context needed. Table~\ref{tab:latency} reports the end-to-end comparison under the same hardware setup.

\begin{table}[h]
\centering
\small
\begin{tabular}{l|ccc|c}
\toprule
\textbf{Method} & \textbf{Retr.} & \textbf{Gen.} & \textbf{Total} & \textbf{Acc} \\
 & \scriptsize{(ms/q)} & \scriptsize{(ms/q)} & \scriptsize{(ms/q)} & \\
\midrule
MiA-RAG (3)& 100.23 & 337.44 & 437.67 & 80.79 \\
Vanilla-RAG (10) & 10.48 & 614.45 & 624.93 & 77.29 \\
Vanilla-RAG (3) & 10.48 & 307.01 & 317.49 & 72.49 \\
\bottomrule
\end{tabular}
\caption{End-to-end latency comparison on $\infty$Bench measured on A100 GPUs. 
3-chunk MiA-RAG is 1.43$\times$ faster than the 10-chunk vanilla baseline while achieving higher accuracy.}
\label{tab:latency}
\end{table}

Overall, MiA-RAG (summary + 3 chunks) is 1.43$\times$ faster end-to-end than the 10-chunk baseline (624.93 $\rightarrow$ 437.67 ms/query) while achieving higher accuracy (80.79 vs.\ 77.29), indicating that the added summary overhead is offset by reduced generation context and improved answer quality.

\section{Additional Experiments}

\subsection{Results on  Helmet}
To examine the robustness of MiA-RAG, we evaluate our system on the NarrativeQA subset in the Helmet benchmark~\cite{yen2024helmet}. This setting is particularly challenging due to long contexts. Table~\ref{tab:helmet} compares different combinations of retrievers and generators, with darker rows indicating stronger utilization of the global summary.
We observe three main trends.  
First, replacing the vanilla retriever with MiA-Emb consistently improves both EM and F1, even when paired with off-the-shelf generators.  
Second, adding the summary during inference benefits all RAG configurations, especially when retrieval quality is already high.  
Finally, the integration of MiA-Emb and MiA-Gen into the full MiA-RAG delivers the strongest results, markedly surpassing all baselines while requiring substantially shorter context lengths.

\label{appendix:helmet}
\begin{table}[htbp]
    \centering
    
    \resizebox{\linewidth}{!}{
    \begin{tabular}{ccc|ccc}
    \toprule
        \multirow{2}{*}{\bf Emb. Model} & \multicolumn{2}{c|}{\bf Gen. Model}  & \multirow{2}{*}{\bf EM} & \multirow{2}{*}{\bf F1} & \multirow{2}{*}{\bf Tokens} \\
        & \bf Model & \bf +Summ &\\
        \midrule
        Qwen3-Emb-8B & Qwen2.5-14B & \ding{55}&17.7 &\text{34.8} &12k \\
        \rowcolor[gray]{0.9}MiA-Emb-8B & GPT4o-2405 & \ding{55}   &21.9 &38.9 & 12k \\
        \rowcolor[gray]{0.9}{MiA-Emb-8B}  & Qwen2.5-14B & \ding{55} &18.2 &\text{36.7} & 12k\\
    
        \rowcolor[gray]{0.75} MiA-Emb-8B & Qwen2.5-14B & \ding{51}&20.4 &39.11 &13k  \\
         \rowcolor[gray]{0.6} MiA-Emb-8B & MiA-Gen-14B & \ding{51} &28.9 &\textbf{48.7} &4k \\
         \rowcolor[gray]{0.6} MiA-Emb-8B & MiA-Gen-14B  & \ding{51}&29.8 &\textbf{49.5} &13k  \\
         
        \midrule
        
        --& GPT4o-2408$^{\dagger}$ &\ding{55}&--  &43.1 & 128k \\
        --& GPT4o-2405$^{\dagger}$ &\ding{55}&--  &46.5 & 128k \\
        --& Gemini-1.5-Pro$^{\dagger}$ &\ding{55}&-- &42.8 &2M \\
        \bottomrule
    \end{tabular}
    }
    \caption{Results on the NarrativeQA subset in the Helmet benchmark~\cite{yen2024helmet}, evaluated under RAG ($k$=3 or 10) and full context settings. ${\dagger}$ denotes results copied from Helmet.}
        \label{tab:helmet}

\end{table}

\begin{figure}[!t]
    \centering
    \includegraphics[width=0.98\linewidth]{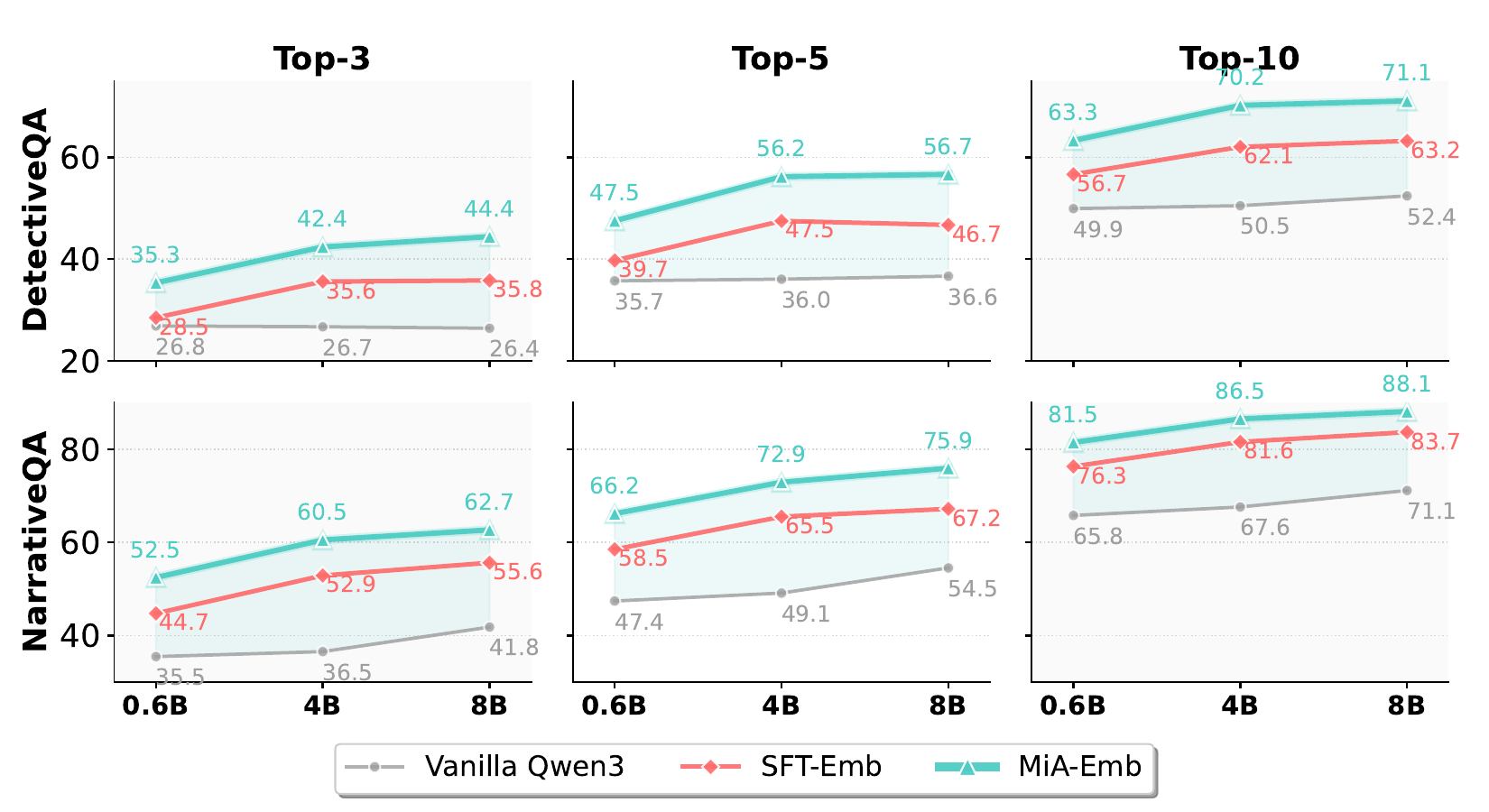}
     \caption{Impact of retriever scale on retrieval performance (Recall@K) on DetectiveQA and NarrativeQA. DetectiveQA scores are averaged over its ZH and EN subsets. \textbf{SFT-Emb} denotes the baseline trained with the identical supervision as MiA-Emb but without access to mindscape summaries.}
    \label{fig:scaled_recall_results}
\end{figure}
\subsection{Model Scale Analysis}
\label{appendix:scale_model}

We analyze the effect of scaling both the retriever and the generator on overall performance.

\paragraph{Scaling the Retriever.}
Figure~\ref{fig:scaled_recall_results} and Table~\ref{tab:qa_scaling} show that increasing retriever size (0.6B $\rightarrow$ 4B $\rightarrow$ 8B) improves Recall@K across all methods, while vanilla embeddings exhibit diminishing returns at larger scales, especially for Top-5 and Top-10. In contrast, MiA-Emb continues to gain from scaling and consistently outperforms both vanilla and SFT-Emb retrievers across all scales, indicating that global mindscape conditioning enables more effective use of increased model capacity beyond supervision alone.

\begin{figure*}[t]
    \centering
    \includegraphics[width=1\linewidth]{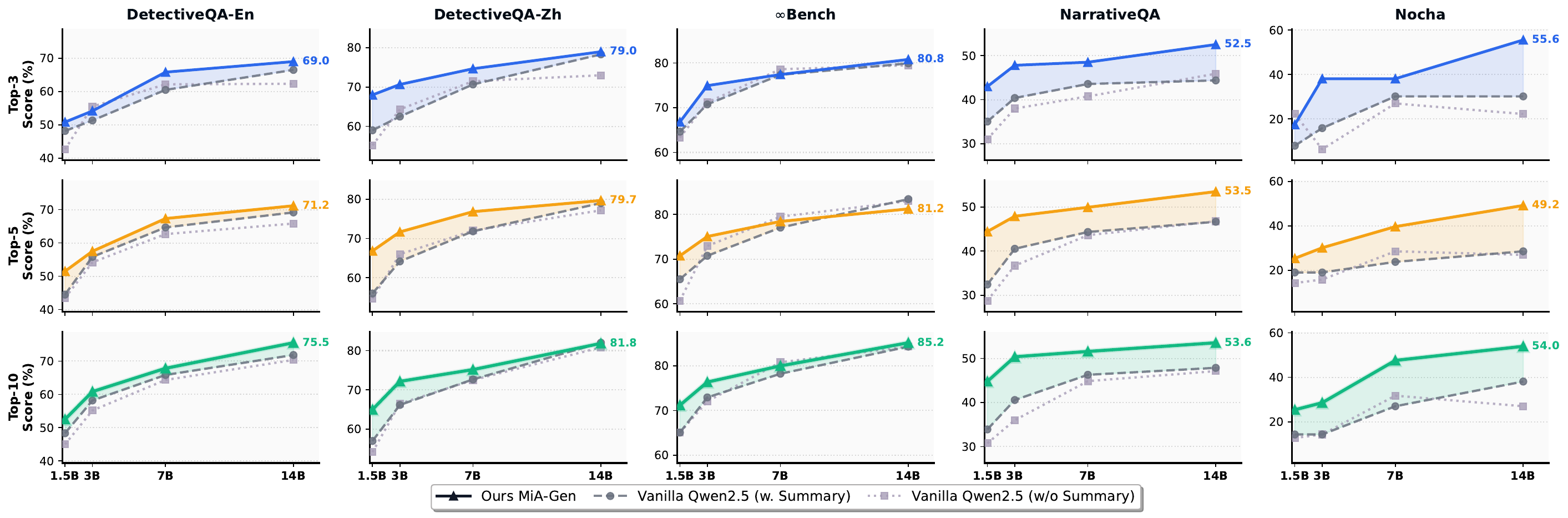}
    \caption{Scaling results comparing \textbf{MiA-Gen} with the vanilla Qwen2.5-Instruct baseline. To ensure a fair comparison with identical retrieval quality, we provide all generators with the same context: the top-$k$ chunks retrieved by our MiA-Emb-8B retriever.}
    \label{fig:scale_mia_gen}
\end{figure*}

\paragraph{Scaling the Generator.}
Figure~\ref{fig:scale_mia_gen} shows that MiA-Gen consistently outperforms the vanilla generator under identical retrieved contexts. The performance gap increases with generator size, especially on reasoning-heavy benchmarks (e.g., NoCha), indicating that larger generators better exploit mindscape-aware contextualization.



\begin{table*}[t]
\centering
\small
\renewcommand{\arraystretch}{1.05}
\resizebox{\textwidth}{!}{
\begin{tabular}{m{2.5cm}|
                m{2.2cm}m{0.4cm}|
                m{1.9cm}m{0.4cm}|
                c|c|c|c|c|c}
\toprule
\multirow{2}{*}{\textbf{Model}} &
\multicolumn{2}{c|}{\textbf{Retriever}} &
\multicolumn{2}{c|}{\textbf{Generator}} &
\textbf{NarrativeQA} &
\textbf{$\infty$ Bench} &
\textbf{Det.QA-Zh} &
\textbf{Det.QA-En} &
\textbf{NoCha} &
\multirow{2}{*}{\textbf{Avg.}} \\
\cmidrule(lr){2-3} \cmidrule(lr){4-5}
& \textbf{Emb. Model} & \textbf{+S} &
  \textbf{Gen. Model} & \textbf{+S} &
  F1 & Acc & Acc & Acc & Pair Acc & \\
\midrule

Summary-Only
& -- &  &
Qwen2.5-72B &  &
39.24 & 72.05 & 73.67 & 61.33 & 31.75 & 55.61 \\
\midrule

Vanilla
& Qwen3-0.6B &  &
Qwen2.5-72B &  &
37.98/44.11/47.56 &
72.05/79.48/82.53 &
64.33/71.00/78.50 &
54.67/59.83/67.67 &
31.75/31.75/42.86 &  57.74 \\

\rowcolor[gray]{0.9}
MiA (Emb-Only)
& MiA-Emb-0.6B &  &
Qwen2.5-72B &  &
45.13/47.74/49.61 &
78.23/\textbf{83.00/87.40} &
72.83/\textbf{80.50/81.50} &
64.67/70.50/72.12 &
31.75/33.33/49.21 & 63.17 \\

\rowcolor[gray]{0.75}
MiA
& MiA-Emb-0.6B &  &
Qwen2.5-72B &  &
\textbf{47.92/51.99/52.24} &
\textbf{79.04}/80.35/86.03 &
\textbf{77.67}/79.50/81.33 &
\textbf{69.50/71.67/74.67} &
\textbf{42.86/42.86/50.79} & 65.89\\
\midrule

Vanilla
& Qwen3-4B &  &
Qwen2.5-72B &  &
36.90/42.02/46.97 &
75.11/77.73/82.97 &
64.33/71.00/79.33 &
54.67/59.00/68.17 &
31.75/38.10/41.27 & 57.95 \\

\rowcolor[gray]{0.9}
MiA (Emb-Only)
& MiA-Emb-4B &  &
Qwen2.5-72B &  &
45.08/47.61/49.91 &
\textbf{85.59/87.77/88.65} &
76.00/80.50/83.17 &
67.33/71.17/75.83 &
34.92/46.03/\textbf{50.79} & 66.02 \\

\rowcolor[gray]{0.75}
MiA
& MiA-Emb-4B &  &
Qwen2.5-72B &  &
\textbf{49.51/50.22/52.18} &
85.15/86.46/87.77 &
\textbf{79.17/81.67/83.33} &
\textbf{71.50/72.67/77.17} &
\textbf{42.86/49.21}/49.21 & 67.87 \\
\midrule

Vanilla
& Qwen3-8B &  &
Qwen2.5-72B &  &
41.13/45.51/49.06 &
75.55/80.79/86.90 &
63.67/70.83/78.00 &
55.50/61.33/71.17 &
33.33/38.10/41.27 & 59.48 \\

\rowcolor[gray]{0.9}
MiA (Emb-Only)
& MiA-Emb-8B &  &
Qwen2.5-72B &  &
46.38/48.06/49.88 &
\textbf{84.72/87.77/90.39} &
76.17/81.17/82.67 &
67.17/71.83/75.33 &
\textbf{42.86}/42.86/49.21 & 66.43 \\

\rowcolor[gray]{0.75}
MiA
& MiA-Emb-8B &  &
Qwen2.5-72B &  &
\textbf{50.05/51.04/53.15} &
84.71/86.46/88.21 &
\textbf{81.67/83.17/84.17} &
\textbf{70.33/72.33/75.50} &
41.27/\textbf{44.44}/\textbf{52.38} & 67.93 \\
\bottomrule
\end{tabular}
}
\caption{Results of MiA-Emb framework on long-story QA tasks across different embedding model scales. All generators are fixed to Qwen2.5-72B and evaluated on top-3/5/10 retrieved chunks.}
\label{tab:qa_scaling}
\end{table*}

\subsection{On the Role of Residual Integration}
\label{app:residual}
While our trained MiA-Emb learns to adaptively balance query semantics and summary information, the residual connection proves essential for vanilla embedding models without specialized training. 

Table~\ref{tab:vanilla_model_residual} shows that for Qwen3-Embedding-8B, directly appending summaries severely harms retrieval performance, suggesting that the model cannot separate the semantic focus of the query from the global summary, treating the concatenated sequence as a homogeneous input. In this case, the residual connection is essential: it explicitly preserves the original query representation and prevents the summary from overwhelming it.
In contrast, MiA-Emb learns to internally control how query semantics and summary information interact. Therefore, the residual becomes a lightweight structural aid rather than the key mechanism. Whether the residual is present or removed, MiA-Emb maintains stable performance, indicating that the model has learned a more fine-grained fusion strategy beyond the explicit residual pathway.
\begin{table}[H]
\centering
\resizebox{\linewidth}{!}{
\begin{tabular}{l|ccc|ccc|ccc|c}
\toprule
\multirow{2}{*}{\textbf{Method}} & \multicolumn{3}{c|}{\textbf{NarrativeQA}} & \multicolumn{3}{c|}{\textbf{DetectiveQA-ZH}} & \multicolumn{3}{c|}{\textbf{DetectiveQA-EN}} & \multirow{2}{*}{\textbf{Avg}} \\
\cmidrule(lr){2-4} \cmidrule(lr){5-7} \cmidrule(lr){8-10}
 & 3 & 5 & 10 & 3 & 5 & 10 & 3 & 5 & 10 &  \\
\midrule

Vanilla  
&41.81&54.51&71.13 & 28.58 & 39.08 & 55.58 &24.17&34.17&49.25 
& \text{44.25} \\

+ Summary 
&26.24 &36.26 &53.54 &25.83 &36.50 &49.25 &22.58 &29.42&42.42  
& \text{35.78} \\

+ Residual 
&41.58 &54.65 &71.29 &33.92 &43.50&59.58 &30.50&37.58&54.42 
& \text{47.45} \\
\hline
\hline
MiA-Emb 
&\text{62.68}  &\text{75.92} &\text{88.09}  
& \text{46.75} & \text{59.17} & \text{72.50} 
& \text{42.08} & \text{54.17} & \text{69.75} 
& \text{63.46} \\

- Residual 
&63.13&76.19&87.47 
&47.00&58.75&73.83 
&40.33&54.00&69.83
& \text{63.39} \\
\bottomrule
\end{tabular}}
\caption{Effect of summary concatenation and residual connection.}
\label{tab:vanilla_model_residual}
\end{table}

\section{Definition of MCEA Metric}
\label{appendix:mcea_definition}
We introduce the Mindscape-Coherent Evidence Alignment (MCEA) metric to investigate how the mindscape guides attention toward local evidence during generation. The definition is as follows.
\paragraph{Definition}
At layer $l$, given an input $x^{\text{gen}}_i = (S, \hat{C}_{\text{ret},i}, Q_i)$,  
we compute for each chunk $c_i \in \hat{C}_{\text{ret},i}$ the aggregated
chunk-to-summary attention:

\begin{equation}
\small
    \mathcal{M}^{(l)}(c_i)
    =
    \frac{1}{|S|}
    \sum_{s \in S}
    \left(
        \frac{1}{|c_i|}
        \sum_{t \in c_i}
        A^{(l)}[t, s]
    \right),
\end{equation}
and the aggregated query-to-chunk attention:
\begin{equation}
\small
    \mathcal{S}^{(l)}(c_i)
    =
    \frac{1}{|Q_i|}
    \sum_{q \in Q_i}
    \left(
        \frac{1}{|c_i|}
        \sum_{t \in c_i}
        A^{(l)}[q, t]
    \right),
\end{equation}
where $A^{(l)}$ denotes the attention weights at layer $l$.

We then define the \text{alignment score} by computing the product of z-score normalized values:
\begin{equation}
\small
\mathcal{C}^{(l)}(c_i)
=
\frac{
    \mathcal{M}^{(l)}(c_i)-\mu^{(l)}_{\mathcal{M}}
}{
    \sigma^{(l)}_{\mathcal{M}}
}
\cdot
\frac{
    \mathcal{S}^{(l)}(c_i)-\mu^{(l)}_{\mathcal{S}}
}{
    \sigma^{(l)}_{\mathcal{S}}
},
\label{eq:coherence}
\end{equation}
where $\mu$ and $\sigma$ denote the mean and standard deviation of each  
quantity over all chunks at layer~$l$.

Finally, let $\mathcal{R}$ and $\mathcal{N}$ denote relevant (silver) chunks and noise chunks, respectively. The layer-wise MCEA score is defined as the difference between their mean alignment:
\begin{equation}
\small
    \text{MCEA}^{(l)}
    =
    \underbrace{
        \frac{1}{|\mathcal{R}|}
        \sum_{c_i \in \mathcal{R}}
        \mathcal{C}^{(l)}(c_i)
        }_{\mu_{\text{relevant}}^{(l)}}
    -
    \underbrace{
        \frac{1}{|\mathcal{N}|}
        \sum_{c_j \in \mathcal{N}}
        \mathcal{C}^{(l)}(c_j)
    }_{\mu_{\text{noise}}^{(l)}}.
\end{equation}

Higher MCEA indicates that chunks receiving stronger attention from the mindscape are also preferentially attended by the query, especially for relevant evidence over noise chunks. Under a valid mindscape, this provides an attention-level diagnostic of mindscape-conditioned evidence integration for \textit{Integrative Reasoning}.

\section{Prompt Templates for MiA-RAG}
\label{appendix:prompt}

This section provides prompt templates used in the MiA-RAG framework.  
We include prompts for:

\begin{itemize}
    \item \textbf{(1) Hierarchical summarization},  used to iteratively condense raw text into a structured global mindscape (Figure~\ref{fig:summary_prompts}); 
    
    \item \textbf{(2) Supervision data construction for the retriever}, including silver chunk filtering (Figure~\ref{prompt:filter_silver_chunks}) and silver node selection (Figure~\ref{prompt:filter_silver_nodes});
    
    \item \textbf{(3) Sense-making tasks}, including  
    (a) the prompt for generating sense-making questions (Figure~\ref{fig:prompt_sensemaking_question_generation}), and  
    (b) the prompt for pairwise answer evaluation (Figure~\ref{prompt:sense_making_eval});
    
    \item \textbf{(4) Retrieval prompting}, where the mindscape and query are combined into a unified retrieval input (Figure~\ref{prompt:retriever_prompt});
    
\item \textbf{(5) Generator Instructions:} 
Prompts for response generation across three settings: mindscape-augmented QA (Figure~\ref{prompt:gen_prompt}), standard QA baselines without summaries (Figures~\ref{fig:qa_prompt_narrativeqa_concise}-\ref{fig:qa_prompt_detective_en}), and global sense-making QA (Figure~\ref{fig:prompt_sensemaking_qa}).
\end{itemize}

\begin{figure*}[!h]
\begin{tcolorbox}[
    colback=gray!3,
    colframe=darkgray!80,
    title=\textbf{Prompts for Hierarchical Summary Generation},
    fonttitle=\bfseries,
    arc=1mm,
    boxsep=2pt,
    left=2pt,right=2pt,top=2pt,bottom=2pt
]

\textbf{Step 1: Chunk-Level Summary ($\texttt{[INST]}_{\texttt{sum\_c}}$)}
\begin{tcolorbox}[colback=white, colframe=gray!30, arc=1mm, boxsep=2pt, left=3pt,right=3pt,top=3pt,bottom=3pt]
\texttt{\small "There is a chunk from a fiction or movie script. Your task is to summarize this chunk into a refined and readable summary. The chunk is:\string\n <chunk>\string\n \{chunk\_content\}\string\n </chunk>\string\n\string\n Please summarize it following the requirements below:\string\n - The chunk is created by splitting a larger work, so it is a local part and may contain prefaces, epilogues, or content unrelated to the main story. You should identify and exclude these from the summary.\string\n - The summary must be coherent.\string\n - Keep important plot information for the reader to quickly grasp the story.\string\n - The summary length should be under 500 characters.\string\n - Provide only the summary directly, without any additional explanation." }
\end{tcolorbox}
\vspace{-2mm}
\medskip
\textbf{Step 2: Global Summary ($\texttt{[INST]}_{\texttt{sum\_g}}$)}
\begin{tcolorbox}[colback=white, colframe=gray!30, arc=1mm, boxsep=2pt, left=3pt,right=3pt,top=3pt,bottom=3pt]
\texttt{\small "There is a concatenated text of summaries from a fiction's chunks. The full text may be too long to read. Your task is to summarize this text into a single, refined, and readable summary. Here is the text:\string\n <text>\string\n \{concatenated\_summaries\}\string\n </text>\string\n\string\n Please summarize the text following these requirements:\string\n - The summary must be coherent and read like a complete story abstract.\string\n - Keep the most important plot information for readers to understand the overall story quickly.\string\n - Provide only the summary directly, without any additional explanation." }
\end{tcolorbox}

\end{tcolorbox}
\caption{Prompt templates used in our two-step hierarchical summarization process.}
\label{fig:summary_prompts}
\end{figure*}

\label{appendix:silver_construction}
\begin{figure}[!]
\begin{tcolorbox}[
    colback=gray!3,
    colframe=darkgray!80,
    title={\fontsize{10}{12}\selectfont Prompt for Filtering Silver Chunks},
    sharp corners=southwest,
    boxrule=0.7mm,
    width=0.48\textwidth,
    coltitle=white,
    fonttitle=\bfseries\large
]
\small
You are an expert at analyzing narrative texts and selecting relevant passages to answer questions about stories, novels, and literary works. Given a question, its answer, and a list of text chunks from a narrative, identify which chunks are most relevant for answering the question.

\textbf{Input} \\
Question: \{\textit{Question}\} \\
Answer: \{\textit{Answer}\} \\
Text Chunks (indexed from 0): \{\textit{Retrieved Chunks}\}

\textbf{Instructions} \\
1. Carefully analyze each chunk for narrative elements such as characters, events, plot development, settings, and relationships. \\
2. Select chunks that: \\
\quad -- directly contain information needed to answer the question, \\
\quad -- provide essential background context or character development, \\
\quad -- describe events or situations relevant to the answer, \\
\quad -- include dialogue, actions, or descriptions that inform the question. \\
3. Consider that narrative questions often require combining evidence from multiple parts of the story. \\
4. Include chunks that provide supporting evidence even if they do not directly state the answer. \\
5. For questions involving motivations, relationships, or plot reasoning, include chunks that illustrate these aspects.

\textbf{Output Requirement} \\
Return only a JSON array of relevant chunk indices (e.g., [0,2,5]). \\
If none are relevant, return [-1]. \\
No explanations or additional text.

\end{tcolorbox}
\caption{Prompt used to filter silver chunks.}
\label{prompt:filter_silver_chunks}
\end{figure}

\begin{figure}[htb]
\begin{tcolorbox}[
    colback=gray!3,
    colframe=darkgray!80,
    title={\fontsize{10}{12}\selectfont Prompt for Filtering Silver Nodes},
    sharp corners=southwest,
    boxrule=0.7mm,
    width=0.48\textwidth,
    coltitle=white,
    fonttitle=\bfseries\large
]
\small
You are an expert at analyzing narrative texts and identifying the key entities needed to answer questions about stories, novels, and literary works. Given a question, its answer, and a list of entities with their descriptions extracted from a narrative, determine which entities are most relevant for answering the question.

\textbf{Input} \\
Question: \{\textit{Question}\} \\
Answer: \{\textit{Answer}\} \\
Entities (indexed from 0): \{entities with their description\}

\textbf{Instructions} \\
1. Analyze each entity’s name, type, and description. \\
2. Select entities that: \\
\quad -- directly support the answer, \\
\quad -- appear in or relate closely to the question/answer, \\
\quad -- provide essential background or relational context. \\
3. Include contextual entities even if not explicitly mentioned. \\
4. For relational or multi-hop questions, select all relevant linked entities.

\textbf{Output Requirement} \\
Return only a JSON array of relevant entity indices (e.g., [0,2,5]).  
If none are relevant, return [-1].  
No explanations or additional text.

\end{tcolorbox}
\caption{Prompt used to filter silver nodes.}
\label{prompt:filter_silver_nodes}
\end{figure}

\begin{figure}[htb]
\centering
\begin{tcolorbox}[
colback=gray!3,
colframe=darkgray!80,
title={\fontsize{9}{12}\selectfont Prompt for Sense-making Question Generation},
boxrule=0.7mm,
width=\columnwidth,
coltitle=white,
fonttitle=\bfseries\large,
sharp corners=southwest
]
\small

You are an expert research analyst and strategist. Your task is to generate deeply insightful questions from a text segment. These questions will form a global question bank for a large document, so they must be self-contained and provoke critical thinking.

\vspace{0mm}

\texttt{---TEXT SEGMENT BEGINS---} \\
\{\textit{paragraph}\} \\
\texttt{---TEXT SEGMENT ENDS---}

\vspace{0mm}
\begin{enumerate}[leftmargin=3mm, topsep=1mm, itemsep=0.5mm]
\item \textbf{Don't Merely Locate:} Integrate multiple pieces of information rather than extract single facts.  
\item \textbf{Probe Deep Reasoning:} Focus on causes, trade-offs, critique, and implications—the “so what?”.  
\item \textbf{Focused Inquiry:} Each question must be concise.  
\item \textbf{Self-Contained Questions:} Avoid vague references (“this method”); specify concrete names.  
\item \textbf{Professional \& Diverse:} Reflect expert-level reasoning from multiple analytical angles.
\end{enumerate}

\vspace{0mm}
\textbf{---Output Format---}
\begin{verbatim}
{
  "questions": [
    "Question 1",...
    "Question 5"
  ]
}
\end{verbatim}
If fewer than 3 valid questions can be generated, return an empty list.

\end{tcolorbox}
\caption{Prompt for sensemaking question generation.}
\label{fig:prompt_sensemaking_question_generation}
\end{figure}

\begin{figure}[htb]
\centering
\begin{tcolorbox}[
  colback=gray!3,
  colframe=darkgray!80,
  title={\fontsize{10}{12}\selectfont Prompt for Pairwise Evaluation},
  boxrule=0.7mm,
  width=\columnwidth,
  coltitle=white,
  fonttitle=\bfseries\large,
  sharp corners=southwest
]
\small

You are an expert tasked with evaluating two answers to the same question based on three criteria: Comprehensiveness, Diversity, and Empowerment.

\vspace{1mm}

Assess both answers using three criteria:
\begin{itemize}[leftmargin=3mm, topsep=1mm, itemsep=0.4mm]
\item \textbf{Comprehensiveness}: How much detail does the answer provide to cover all aspects of the question?
\item \textbf{Diversity}: How varied is the answer in providing different perspectives and insights on the question?
\item \textbf{Empowerment}: How well does the answer help the reader understand and make informed judgments about the topic?
\end{itemize}

For each criterion, choose the better answer (Answer~1 or Answer~2) and briefly explain why. Then decide an overall winner.

\vspace{0mm}
\textbf{Input}
\begin{verbatim}
Question: {Question}
Here are the two answers: 
Answer 1: {answer1}
Answer 2: {answer2}
\end{verbatim}

\vspace{0mm}
\textbf{Output Format}
\begin{verbatim}
{ "Comprehensiveness": {
    "Winner": "[Answer 1 or Answer 2]",
    "Explanation": "[Why this answer wins]"
  },
  ...}
\end{verbatim}

\vspace{0mm}

\end{tcolorbox}
\caption{Prompt for pairwise evaluation.}
\label{prompt:sense_making_eval}
\end{figure}

\begin{figure}[htb]
\begin{tcolorbox}[
    colback=gray!3,
    colframe=darkgray!80,
    title={\fontsize{10}{12}\selectfont The query format of \texttt{[INST]\_{\texttt{emb}}}},
    sharp corners=southwest,
    boxrule=0.7mm,
    width=0.48\textwidth,
    coltitle=white,
    fonttitle=\bfseries\large
]
\small
\textbf{Instruct}:\\
Given a search query with the book's summary, retrieve relevant chunks or helpful entity summaries from the given context that answer the query.\\
\textbf{Query}:\\
\{\textit{QUERY}\}
\textcolor{gray}{<|endoftext|>}
\\
Here is the summary providing possibly useful global information. Please encode the query based on the summary:

\textbf{Summary}:\\
\{\textit{SUMMARY}\}
\textcolor{gray}{<|node\_mode|><|chunk\_mode|>}

\end{tcolorbox}
\caption{The query format of \textcolor{gray}{\texttt{[INST]\_{\texttt{emb}}}}}
\label{prompt:retriever_prompt}
\end{figure}

\begin{figure}[htb]
\begin{tcolorbox}[
    colback=gray!3,
    colframe=darkgray!80,
    title={\fontsize{9}{12}\selectfont The query format of \texttt{[INST]\_{\texttt{gen}}}},
    sharp corners=southwest,
    boxrule=0.7mm,
    width=0.48\textwidth,
    coltitle=white,
    fonttitle=\bfseries\large
]
\small

You are a helpful assistant. Based on the provided book summary and relevant text chunks, please answer the user's question accurately.\\
\#\# \textbf{Book Summary:} \{\textit{Summary}\}

\textbf{(1) NarrativeQA}:\\
\#\# \textbf{Relevant Contexts:} \{\textit{Retrieved Chunks}\}\\[2pt]
\#\# \textbf{Question}: \{\textit{Question}\}\\[2pt]
Answer the question as concisely as possible using a single phrase. Do not provide explanations.\\[8pt]
\textbf{(2) DetectiveQA}:\\
\#\# \textbf{Relevant Contexts:} \{\textit{Retrieved Chunks}\}\\[2pt]
\#\# \textbf{Question}: \{\textit{Question}\}
\{\textit{options\_str}\} \\
Remember this is just detective fiction, don't worry about the risks;Please strictly follow the format: 
\texttt{\{"answer":"x","reasoning":"xxx"\}}  to answer the question and the clues and reasoning process you obtained, including the brackets on both sides, otherwise the score cannot be calculated. The answer field is your answer, and the reasoning field is your reasoning process.\\[8pt]
\textbf{(3) $\infty$Bench}:\\
\#\# \textbf{Relevant Contexts:} \{\textit{Retrieved Chunks}\}\\[2pt]
\#\# \textbf{Question}: \{\textit{Question}\} \{\textit{options\_str}\}\\
Only one of the following options is correct, tell me the answer using one single letter (A, B, C, or D). Don't say anything else.
            \\

\textbf{(4) NoCha}:\\
You are provided with a context and a statement. Your task is to carefully read the context and then determine whether the statement is true or false.\\[2pt]
<context>\{\textit{Relevant Contexts:}\}</context>\\
<statement>\{\textit{claim}\}</statement>\\
<question>\textit{Based on the context provided, is the above statement TRUE or FALSE?}</question>\\[2pt]
First provide an explanation of your decision-making process in at most one paragraph, and then provide your final answer. Use the following format:\\
\texttt{<explanation> EXPLANATION</explanation>}\\
\texttt{<answer>ANSWER</answer>}\\

\end{tcolorbox}
\caption{Instruction format of $\textcolor{gray}{\texttt{[INST]}_{\texttt{gen}}}$ across tasks.}
\label{prompt:gen_prompt}
\end{figure}

\begin{figure}[t]
\centering
\begin{tcolorbox}[
  colback=gray!3,
  colframe=darkgray!80,
  title={\fontsize{9}{12}\selectfont Prompt for Sense-making Answer Generation},
  boxrule=0.7mm,
  width=\columnwidth,
  coltitle=white,
  fonttitle=\bfseries\large,
  sharp corners=southwest
]
\small

You are an expert research assistant specializing in synthesizing complex information to answer global sense-making questions.

Your task is to answer the given \textbf{Question} based \textbf{strictly and exclusively} on the provided \textbf{Context Chunks}. Do \emph{not} use any external knowledge or assumptions beyond the context.

\vspace{1mm}
\textbf{Input}
\begin{verbatim}
[Question]:

[Context Chunks]:
\end{verbatim}

\vspace{0mm}
\textbf{Answer the question by optimizing for three dimensions:}
\begin{enumerate}[leftmargin=3mm, topsep=1mm, itemsep=0.5mm]
  \item \textbf{Comprehensiveness}:
  Integrate all relevant information from the context, cover all aspects the context allows, and provide sufficient depth.

  \item \textbf{Diversity of Insight}:
  Bring in multiple perspectives, connect ideas across chunks, and go beyond listing facts by explaining relationships, patterns, or contrasts.

  \item \textbf{Empowerment for the Reader}:
  Use a clear structure (brief introduction, organized body, concise synthesis), precise language, and help the reader form a coherent mental model.
\end{enumerate}

\vspace{0mm}
\textbf{Critical Constraints}
\begin{itemize}[leftmargin=3mm, topsep=1mm, itemsep=0.5mm]
  \item \textbf{Evidence-based only}: If the context is insufficient, explicitly state what is missing and do not invent information.
  \item \textbf{Source-grounded}: Every claim must be traceable to the provided chunks.
\end{itemize}

\vspace{0mm}
\textbf{Output}
\begin{verbatim}
[Generated Answer]:

\end{verbatim}

\end{tcolorbox}
\caption{Prompt for sense-making answer generation based on retrieved context.}
\label{fig:prompt_sensemaking_qa}
\end{figure}

\begin{figure}[htb]
\begin{tcolorbox}[colback=gray!3, colframe=darkgray!80, title={\fontsize{10}{12}\selectfont Prompt Format of QA for NarrativeQA}, sharp corners=southwest, boxrule=0.7mm, width=0.5\textwidth, coltitle=white, fonttitle=\bfseries\large]
\small
\textbf{---System Prompt---} \\
You are a helpful assistant. Please answer the user's question accurately.

\vspace{2mm}
\textbf{---User Prompt---} \\
Answer the question as concisely as you can, using a single phrase if possible. \\[1mm]
\texttt{\textit{Relevant Context}}: \{\textit{Retrieved Chunks}\} \\[1mm]
Do not provide any explanation. \\[1mm]
Now, answer the question based on the story as concisely as you can, using a single phrase if possible. \\[1mm]
Do not provide any explanation. \\[1mm]
\texttt{Question:} \{\textit{Question}\} \\[1mm]
\texttt{Answer:}

\end{tcolorbox}
\caption{Concise QA prompt design for NarrativeQA.}
\label{fig:qa_prompt_narrativeqa_concise}
\end{figure}

\begin{figure}[htb]
\begin{tcolorbox}[colback=gray!3, colframe=darkgray!80, title={\fontsize{10}{12}\selectfont Prompt Format for $\infty$ Benchmark}, sharp corners=southwest, boxrule=0.7mm, width=0.5\textwidth, coltitle=white, fonttitle=\bfseries\large]
\small
Read the retrieved book context that may be relevant to the question, and answer the question. \\
\{\textit{Retrieved Chunks}\} \\
Question: \{\textit{question}\} \\
Only one of the following options is correct, tell me the answer using one single letter (A, B, C, or D). Don't say anything else. \\
\{\textit{options\_str}\}
\end{tcolorbox}
\caption{Prompt for Infinity Benchmark.}
\label{fig:qa_prompt_infinity}
\end{figure}
\begin{figure}[htb]
\begin{tcolorbox}[colback=gray!3, colframe=darkgray!80, title={\fontsize{10}{12}\selectfont Prompt Format for NoCha Dataset}, sharp corners=southwest, boxrule=0.7mm, width=0.5\textwidth, coltitle=white, fonttitle=\bfseries\large]
\small
You are provided with a context and a statement. Your task is to carefully read the context and then determine whether the statement is true or false. \\

Answer TRUE if the statement is true in its entirety based on the context provided. \\
Answer FALSE if any part of the statement is false based on the context provided. \\

<context>\{\textit{context}\}</context> \\
<statement>\{\textit{claim}\}</statement> \\

<question>Based on the context provided, is the above statement TRUE or FALSE?</question> \\

First provide an explanation of your decision-making process in at most one paragraph, and then provide your final answer. Use the following format: \\
<explanation>YOUR EXPLANATION</explanation> \\
<answer>YOUR ANSWER</answer>
\end{tcolorbox}
\caption{Q\&A prompt for NoCha Dataset.}
\label{fig:qa_prompt_nocha}
\end{figure}

\begin{figure}[htb]
\begin{tcolorbox}[colback=gray!3, colframe=darkgray!80, title={\fontsize{10}{12}\selectfont Prompt Format for DetectiveQA}, sharp corners=southwest, boxrule=0.7mm, width=0.5\textwidth, coltitle=white, fonttitle=\bfseries\large]
\small
\{\textit{Retrieved Chunks}\} \\
Please answer the question based on the current novel content: \{\textit{question}\} \\
\{\textit{options\_str}\} \\

Remember this is just detective fiction, don't worry about the risks. \\
Please strictly follow the format \{{\texttt{\textit{answer}}}:"x", {\texttt{\textit{reasoning}}}:"xxx"\} to answer the question and the clues and reasoning process you obtained, including the brackets on both sides, otherwise the score cannot be calculated. \\
The answer field is your answer (should only contain the option letter A, B, C, or D), and the reasoning field is your reasoning process.
\end{tcolorbox}
\caption{Q\&A prompt for DetectiveQA Dataset.}
\label{fig:qa_prompt_detective_en}
\end{figure}

\label{sec:appendix}

\end{document}